\documentclass[10pt,journal,compsoc]{IEEEtran}
\ifCLASSOPTIONcompsoc
  \usepackage[nocompress]{cite}
\else
  \usepackage{cite}
\fi

\ifCLASSINFOpdf
\else
\fi

\usepackage{booktabs}
\usepackage{paralist} 
\usepackage{mathrsfs,amsfonts,amsmath,amsthm,amssymb}
\usepackage{multicol}
\usepackage{subfigure}
\usepackage{graphicx}
\usepackage{multirow}
\usepackage{subfigure}
\usepackage{algorithm}
\usepackage{algorithmic}
\usepackage{setspace}
\usepackage{siunitx}
\usepackage[most]{tcolorbox}
\usepackage{adjustbox}
\usepackage{balance}

\usepackage{tikz}    
\usepackage{tikz-qtree}

\usepackage{CJKutf8}
\usetikzlibrary{trees}
\usetikzlibrary{external}
\usetikzlibrary{shapes.geometric, arrows, shadows, calc,positioning,external}
\usetikzlibrary{arrows.meta}
\usetikzlibrary{shapes}
\theoremstyle{definition}

\theoremstyle{definition}

\theoremstyle{definition}

\theoremstyle{definition}

\usepackage{color}

\usepackage{url}

\begin{document}
%
\title{Natural Language Interfaces for Tabular Data Querying and Visualization: A Survey}

\author{Weixu Zhang, Yifei Wang, Yuanfeng Song, Victor Junqiu Wei, Yuxing Tian, Yiyan Qi, Jonathan H. Chan, Raymond Chi-Wing Wong, and Haiqin~Yang,~\IEEEmembership{Senior Member,~IEEE} 
\thanks{
W.~Zhang (Xi'an Jiaotong University, email:~weixu\_zhang@stu.xjtu.edu.cn), Y.~Wang (University of Toronto, email:~yifeii.wang@mail.utoronto.ca), and Y.~Tian (Xidian University, email:~tianyxxx@stu.xidian.edu.cn) are interns at International Digital Economy Academy (IDEA), Shenzhen, China}
\thanks{Y.~Song is with WeBank Co., Ltd., Shenzhen, China. Email:~yfsong@webank.com}
\thanks{V.~J.~Wei and R.~C.~Wong are with Department of Computer Science and Engineering, Hong Kong University of Science and Technology (HKUST), Hong Kong.  Email:~\{victorwei,raywong\}@cse.ust.hk}
\thanks{Y.~Qi is with IDEA, Shenzhen, China.  Email:~qiyiyan@idea.edu.cn}
\thanks{J.~H.~Chan is with Innovative Cognitive Computing (IC2) Research Center at School of Information Technology, King Mongkut's University of Technology Thonburi. Email:~jonathan@sit.kmutt.ac.th}
\thanks{H.~Yang (corresponding author) is affiliated with IDEA. Email: hqyang@ieee.org}

}

\markboth{IEEE Transactions on Knowledge and Data Engineering,~Vol.~xx, No.~xx, August~20xx}%
{Shell \MakeLowercase{\textit{et al.}}: Bare Advanced Demo of IEEEtran.cls for IEEE Computer Society Journals}
%



\IEEEtitleabstractindextext{%
\begin{abstract}

The emergence of natural language processing has revolutionized the way users interact with tabular data, enabling a shift from traditional query languages and manual plotting to more intuitive, language-based interfaces. The rise of large language models (LLMs) such as ChatGPT and its successors has further advanced this field, opening new avenues for natural language processing techniques. This survey presents a comprehensive overview of natural language interfaces for tabular data querying and visualization, which allow users to interact with data using natural language queries. We introduce the fundamental concepts and techniques underlying these interfaces with a particular emphasis on semantic parsing, the key technology facilitating the translation from natural language to SQL queries or data visualization commands. We then delve into the recent advancements in Text-to-SQL and Text-to-Vis problems from the perspectives of datasets, methodologies, metrics, and system designs. This includes a deep dive into the influence of LLMs, highlighting their strengths, limitations, and potential for future improvements. Through this survey, we aim to provide a roadmap for researchers and practitioners interested in developing and applying natural language interfaces for data interaction in the era of large language models. 

\end{abstract}

}

\maketitle

\IEEEdisplaynontitleabstractindextext


%
\IEEEpeerreviewmaketitle

\section{Introduction}\label{sec:intro}

\IEEEPARstart{T}{abular}, or structured, data form the backbone of many fields in today's digital age, including business, healthcare, and scientific research~\cite{DBLP:conf/iui/PopescuEK03,DBLP:journals/pvldb/LiJ14}. 
However, the ability to interact effectively and efficiently with vast amounts of structured data to extract valuable insights remains a crucial challenge.  Traditional methods of interaction, such as querying with structured query languages or manual plotting a visualization, often require a significant degree of technical expertise, thereby limiting their accessibility to a wider user base~\cite{DBLP:journals/nle/AndroutsopoulosRT95}.

With the emergence of natural language processing technologies, the way we interact with structured data is beginning to shift. These technologies enable the development of Natural Language Interfaces (NLIs), making tabular data querying and visualization more intuitive and accessible. Through these interfaces, users can extract information from databases or generate visual representations of data using natural language queries and commands~\cite{DBLP:journals/tvcg/ShenSLYHZTW23,DBLP:conf/coling/IacobBATHR20}. This shift towards language-based interfaces marks a significant stride towards simplifying data interaction, making it more user-friendly and accessible to non-technical users.

The foundational technologies powering these language-based interfaces are rooted in semantic parsing tasks, which transform natural language queries into formal representations tailored for execution on structured databases~\cite{DBLP:conf/akbc/KamathD19}. While various formal languages and functional representations have been introduced for this purpose, such as Prolog, Datalog, and FunQL, two are particularly dominant in tabular data interaction: SQL for data querying and visualization specifications for data visualization. SQL has been the de facto standard for querying relational databases for decades, offering comprehensive operations to retrieve and manipulate data. Visualization specifications provide a structured way to represent complex visualizations, making them an integral part of the data visualization process.
Notably, data querying and visualization are two of the most critical technical directions in tabular data interaction, and they are often intertwined in practical applications. Querying is frequently a sub-step in the visualization process, as users need to first retrieve the relevant data before visualizing it. In real-world scenarios, such as generating data reports, these two tasks are commonly used together to extract insights and present them in a visually appealing and informative manner. For example, a business analyst might use natural language to query a sales database for "total revenue by product category in the last quarter", and then request a "bar chart showing the revenue breakdown" to include in their quarterly report. Given their importance, widespread use, and interconnected nature, this survey will focus on these two types of representations, delving deep into the challenges and advancements in the tasks of translating natural language into SQL (Text-to-SQL) and visualization specifications (Text-to-Vis).


The development of these two semantic parsing tasks has evolved significantly over the years, driven by advancements in machine learning and natural language processing techniques. Early approaches often rely on rule-based or template-based systems~\cite{DBLP:journals/vldb/AffolterSB19,DBLP:conf/akbc/KamathD19} and shallow parsing techniques. However, these methods struggle with complex queries and visualizations and are sensitive to the specific phrasing of the user's input.
Introducing neural networks and deep learning methods brings about a significant leap in performance. These methods, often based on sequence-to-sequence models~\cite{DBLP:journals/vldb/KatsogiannisMeimarakisK23}, can capture more complex patterns in the data and are more robust to variations in the input. However, they still require substantial amounts of training data and struggle with out-of-domain queries~\cite{DBLP:conf/coling/Deng0022}.
The rise of Pretrained Language Models (PLMs), such as BERT~\cite{DBLP:conf/naacl/DevlinCLT19}, T5~\cite{DBLP:journals/jmlr/RaffelSRLNMZLL20}, GPT~\cite{DBLP:conf/iclr/PatelLRCRC23}, marks a turning point in the field. With their ability to leverage pre-training on vast amounts of text data, PLMs have shown remarkable success in a wide range of natural language processing tasks, including Text-to-SQL and Text-to-Vis. Recently, the advent of Large Language Models (LLMs) such as ChatGPT and the exploration of prompt engineering techniques have opened new avenues for developing more effective and user-friendly NLIs for data interaction.


The interdisciplinary research on NLIs for tabular data querying and visualization incorporates multiple research aspects, such as natural language processing and data mining, with advancements often following diverse and distinct trajectories. Despite its increasing importance, no single study has comprehensively reviewed the problem of semantic parsing for both querying and visualization tasks in a systematic and unified manner.
Prior surveys have primarily focused on early approaches and subsequent deep learning developments in querying~\cite{DBLP:conf/coling/IacobBATHR20,DBLP:journals/vldb/KatsogiannisMeimarakisK23,DBLP:conf/coling/Deng0022} and visualization~\cite{DBLP:journals/vldb/AffolterSB19,DBLP:journals/tvcg/ShenSLYHZTW23} separately, without offering a consolidated view of these intertwined domains. Furthermore, to the best of our knowledge, no existing surveys cover the recent advancements by LLMs in these areas. The profound influence of LLMs on NLIs for data querying and visualization is a rapidly growing area that requires more attention and exploration.

This survey aims to fill these gaps by offering a comprehensive overview of NLIs for tabular data querying and visualization, emphasizing their interconnected nature and the impact of LLMs. By providing a holistic perspective, we will thoroughly analyze the relationship between these two critical tasks, exploring how they can be unified from the perspective of semantic parsing and how advances in one task can inform the other. We will also examine the practical applications where these tasks are used in conjunction and how NLIs can streamline data-driven workflows. Furthermore, we will provide an in-depth review of the latest LLM-based approaches, discussing how they have pushed the state-of-the-art and opened up new possibilities for more powerful and user-friendly NLIs.
We source references from key journals and conferences over the past two decades, spanning Natural Language Processing, Human-Computer Interaction, Data Mining, and Visualization. Our search is guided by terms such as "Natural Language Interface", "Visualization", and "Text-to-SQL", and we also explore cited publications to capture foundational contributions.
Through this survey, we aim to address a set of critical research questions:

$\bullet$ How have NLIs for tabular data querying and visualization evolved over time?

$\bullet$ What is the relationship between these two tasks, and how can they be unified from the perspective of semantic parsing?

$\bullet$ How have recent advancements, especially LLMs, influenced the field?

$\bullet$ What are the inherent strengths and weaknesses of existing methods?

By drawing upon an extensive literature review and analysis, we aim to provide informed and insightful answers to these questions. We will delve into functional representations, datasets, evaluation metrics, and system architectures, particularly emphasizing the influence of LLMs. Our goal is to present a clear and succinct overview of the current state of the art, emphasizing existing approaches' strengths and limitations while exploring potential avenues for future enhancements.

\section{Background and Framework}
\label{section:background}
\subsection{Context}

\begin{figure}[t]
\centering
\includegraphics[width=\columnwidth]{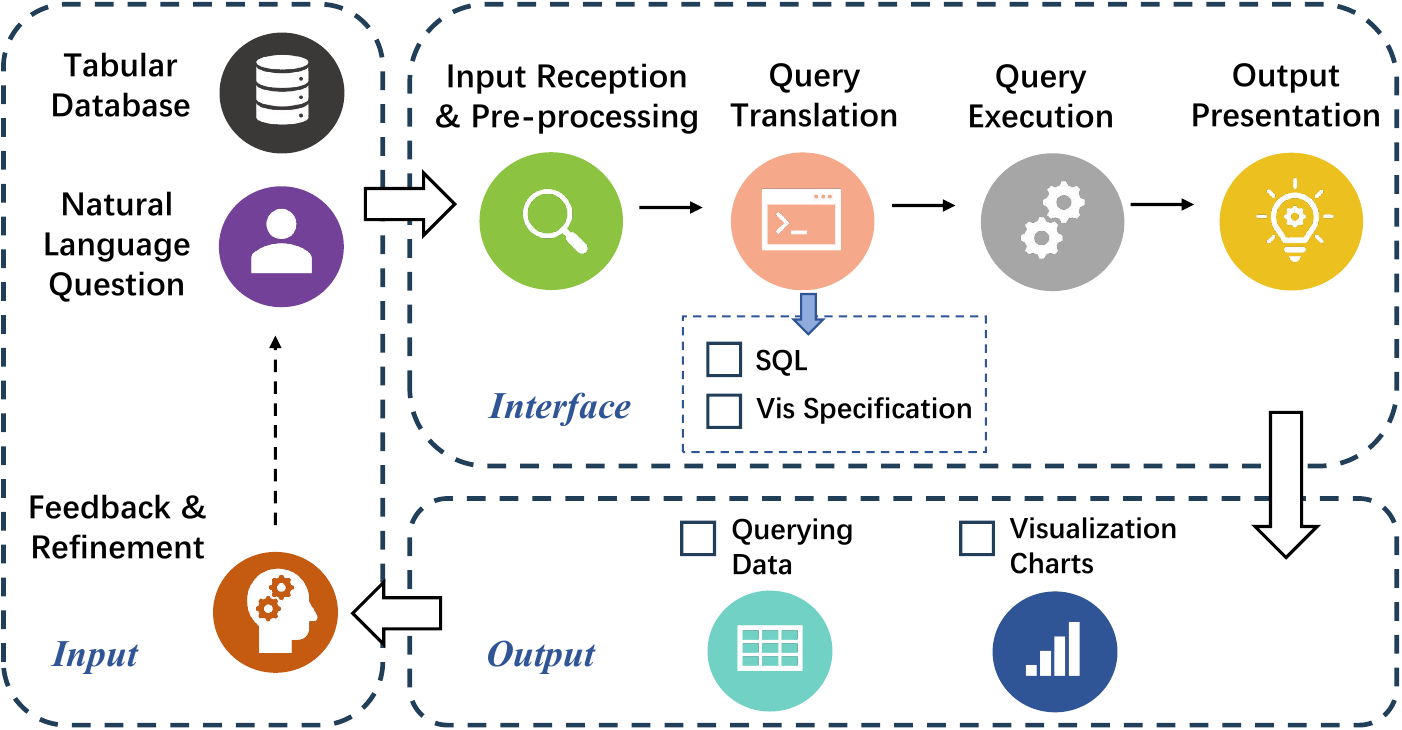}
\caption{Schematic representation of natural language interfaces for tabular data querying and visualization}
\label{fig:problem}
\end{figure}

\begin{figure*}[htbp]
\centering
\includegraphics[width=0.83\textwidth]{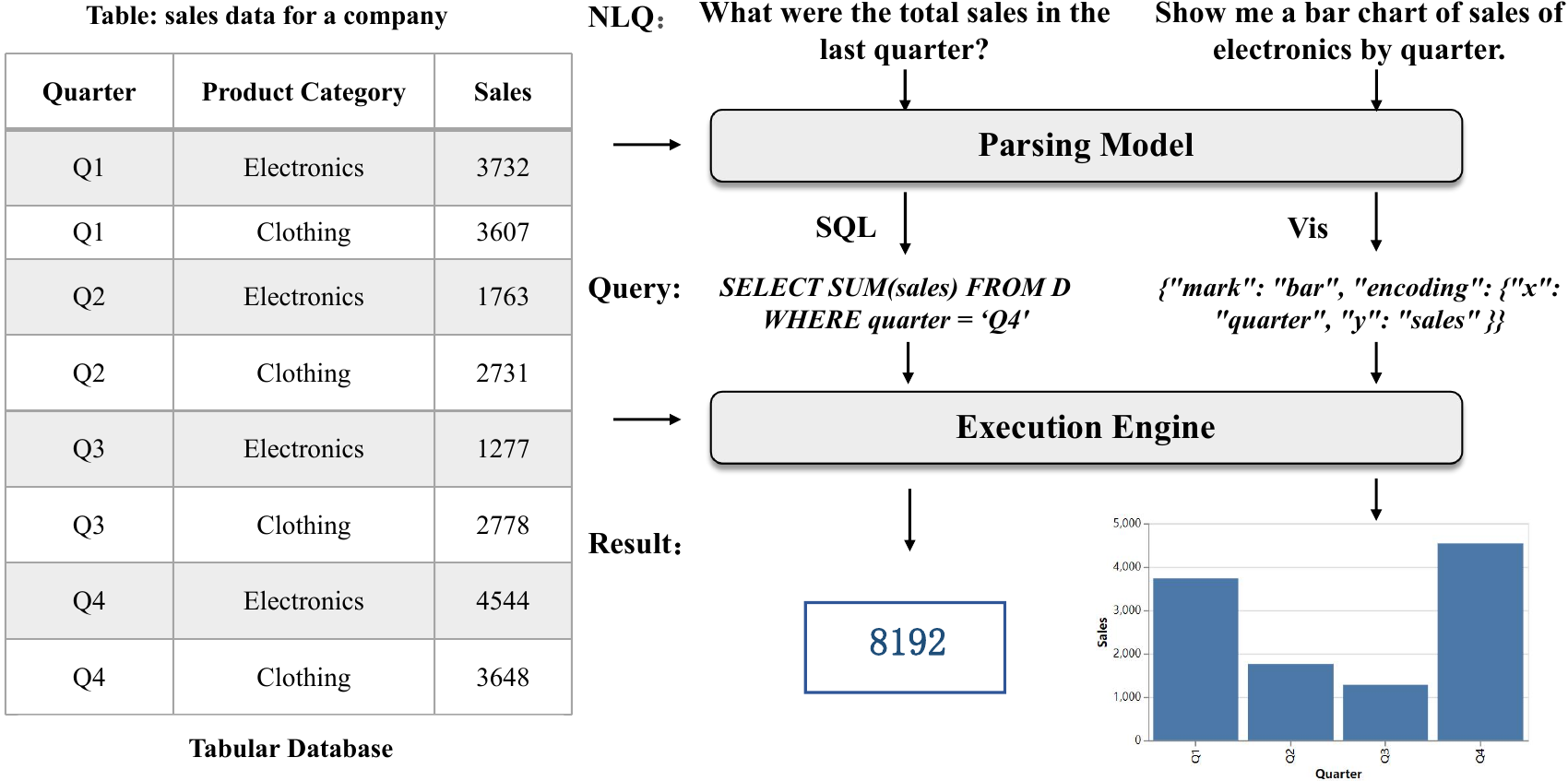}
\caption{Example of the process of translating natural language queries to SQL and visualization specifications on sales data. A textual query about quarterly sales is parsed into an SQL command to fetch numerical data, and a request for a sales visualization is transformed into the corresponding bar chart specification.}
\label{fig:example}
\end{figure*}

The need for Natural Language Interfaces to process tabular data arises from the growing importance of data-driven decision-making across various industries, making it a crucial ability to interact with data efficiently and intuitively. Natural language interfaces simplify access to valuable insights by enabling a wider user base, including those without technical expertise, to query and visualize structured data~\cite{DBLP:journals/nle/AndroutsopoulosRT95}.

Figure~\ref{fig:problem} shows the workflow of natural language interfaces for tabular data querying and visualization, where the user provides input in the form of a natural language question targeting a specific structured database. The interface pre-processes this input, translating it into functional representations, such as SQL queries for data extraction or visualization specifications for chart generation. Executing the SQL queries retrieves relevant data from the database, and the visualization specifications produce corresponding charts. The resulting output, whether raw data or visuals, is then presented to the user, who can provide feedback or further refine their query. This streamlined process enables users to extract data insights and generate visuals without diving into the complexities of databases or visualization tools merely by posing their questions.

The practical application of natural language interfaces for tabular data querying and visualization is exemplified in several existing tools. Microsoft's Power BI~\cite{DBLP:journals/vldb/QinLTL20}, for instance, includes a feature called Q\&A which allows users to ask natural language questions about their data and receive answers in the form of charts or tables. This feature leverages advanced natural language processing to understand the question and generate appropriate visualizations, thereby simplifying the process of data exploration for users.
Similarly, Tableau~\cite{DBLP:conf/iv/Stockl22}, a popular data visualization tool, includes a feature named Ask Data. Users can type a question, and the system generates an answer through a data visualization. These applications underscore the potential and impact of natural language interfaces in enhancing the accessibility and usability of data interaction.

\subsection{Problem Definition}

In the context of natural language interfaces for tabular data, the central problem is to parse a natural language query into a functional representation that can be executed on a structured database.

Formally, given the input $x=\{q, s\}$ with a natural language query $q$ and database schema $s$ containing tables $T=\{t_i\}_{i=1}^{|T|}$ and columns $C=\{q_i\}_{i=1}^{|C|}$ for each table $t_i \in T$, the task of the semantic parser $P$ is to translate $q$ into a functional expression $e$. Once the functional expression $e$ is generated, it can be executed on the structured database $D$ by an execution engine $E$ to produce a result $r$, represented as:
$$
E(e, D) \rightarrow r
$$

The functional expression $e$ and result $r$ vary depending on the specific task:

$\bullet$ \textit{Text-to-SQL.} The functional expression $e$ is an SQL query that manages and queries data held in relational databases~\cite{DBLP:journals/corr/abs-1709-00103}. The result $r$ obtained through the execution of the SQL query is a piece or set of precise data.

$\bullet$ \textit{Text-to-Vis.} The functional expression $e$ is a visualization specification (e.g., Vega-Lite, D3.js) that determines how data should be presented visually, often in the form of charts, graphs, or other graphical elements~\cite{DBLP:journals/tvcg/ShenSLYHZTW23}. The result $r$ obtained through the execution of the visualization specification is a graphical representation such as a pie chart, bar graph, or scatter plot.

$\bullet$ \textit{Prolog and Datalog.} The functional expression $e$ is a set of rules and facts that define relationships and logic for operations on the data. The result $r$ can be in any structure representing the outcome of the query.

$\bullet$ \textit{FunQL.} The functional expression $e$ is an intermediate query language that maps natural language constructs into structured queries, emphasizing the relationships between entities. The result $r$ is a database query language.

The overall process can be seen as a translation from a natural language query $q$ to a result $r$, facilitated by a semantic parser $P$ and an execution engine $E$.

\subsection{Framework}

\begin{figure*}[htbp]
    \centering
    \begingroup
    \tikzset{every picture/.style={scale=0.5}}
    \begin{tikzpicture}[
    node distance=1cm and 1cm,
    every node/.style={draw, rectangle, rounded corners, minimum width=3cm, minimum height=0.6cm}
]

\node[draw, rectangle, rounded corners, font={\normalsize\bfseries}, minimum width=5cm, minimum height=0.8cm] (root) {Natural Language Interfaces for Tabular Data Querying and Visualization};

\newcommand\ChildNode[4]{ 
    \node[below=1cm of #1.west, anchor=west, xshift=0] (#2) {#2};
    \node[below=of #2.west, anchor=west, xshift=0] (#3) {#3};
    \node[below=of #3.west, anchor=west, xshift=0] (#4) {#4};
    \draw ([xshift=-0.7cm]#1.west) -- ++(0,-6cm) ;
    \draw ([xshift=-0.7cm]#1.west)  -- ++ (0.7cm,0) ;
    \draw[-Latex] ([xshift=-0.7cm]#2.west)  -- ++ (0.7cm,0) ;
    \draw[-Latex] ([xshift=-0.7cm]#3.west)  -- ++ (0.7cm,0) ;
    \draw[-Latex] ([xshift=-0.7cm]#4.west)  -- ++ (0.7cm,0) ;
}
\newcommand\ChildNodes[5]{
    \node[below=0.9cm of #1.west, anchor=west, xshift=0] (#2) {#2};
    \node[below=0.9cmof #2.west, anchor=west, xshift=0] (#3) {#3};
    \node[below=0.9cmof #3.west, anchor=west, xshift=0] (#4) {#4};
    \node[below=0.9cmof #4.west, anchor=west, xshift=0] (#5) {#5};
    \draw ([xshift=-0.7cm]#1.west) -- ++(0,-7.2cm) ;
    \draw ([xshift=-0.7cm]#1.west)  -- ++ (0.7cm,0) ;
    \draw[-Latex] ([xshift=-0.7cm]#2.west)  -- ++ (0.7cm,0) ;
    \draw[-Latex] ([xshift=-0.7cm]#3.west)  -- ++ (0.7cm,0) ;
    \draw[-Latex] ([xshift=-0.7cm]#4.west)  -- ++ (0.7cm,0) ;
    \draw[-Latex] ([xshift=-0.7cm]#5.west)  -- ++ (0.7cm,0) ;
}
\node[font={\bfseries},below=0.8cm of root,anchor=north, xshift=-7.3cm] (functional) {Formal Language};
\ChildNode{functional}{SQL Query}{Vis Specification}{Others}

\node[font={\bfseries},below=0.8cm of root,anchor=north, xshift=-3.65cm] (dataset) {Dataset};
\ChildNode{dataset}{Single Domain}{Cross Domain}{Others}

\node[font={\bfseries},below=0.8cm of root,anchor=north, xshift=0] (approach) {Approach};
\ChildNode{approach}{Traditional}{Neural Network}{Foundation Model}

\node[font={\bfseries},below=0.8cm of root,anchor=north, xshift=3.65cm] (evaluation) {Evaluation Metric};
\ChildNode{evaluation}{String-based}{Execution-based}{Manual Evaluation}

\node[font={\bfseries},below=0.8cm of root,anchor=north, xshift=7.3cm] (design) {System Design};
\ChildNodes{design}{Rule-based}{Parsing-based}{Multi-stage}{End-to-end}

\draw[-Latex] (root) -- ++(0,-1.5cm) -| (functional);
\draw[-Latex] (root) -- ++(0,-1.5cm) -| (dataset);
\draw[-Latex] (root) -- ++(0,-1.5cm) -| (approach);
\draw[-Latex] (root) -- ++(0,-1.5cm) -| (evaluation);
\draw[-Latex] (root) -- ++(0,-1.5cm) -| (design);

\end{tikzpicture}
    \endgroup
    \caption{Framework for Natural Language Interfaces in Tabular Data Querying and Visualization. The technology framework comprises various pivotal components: functional representation, dataset, approach, evaluation metric, and system design. }
    \label{fig:framework}
\end{figure*}
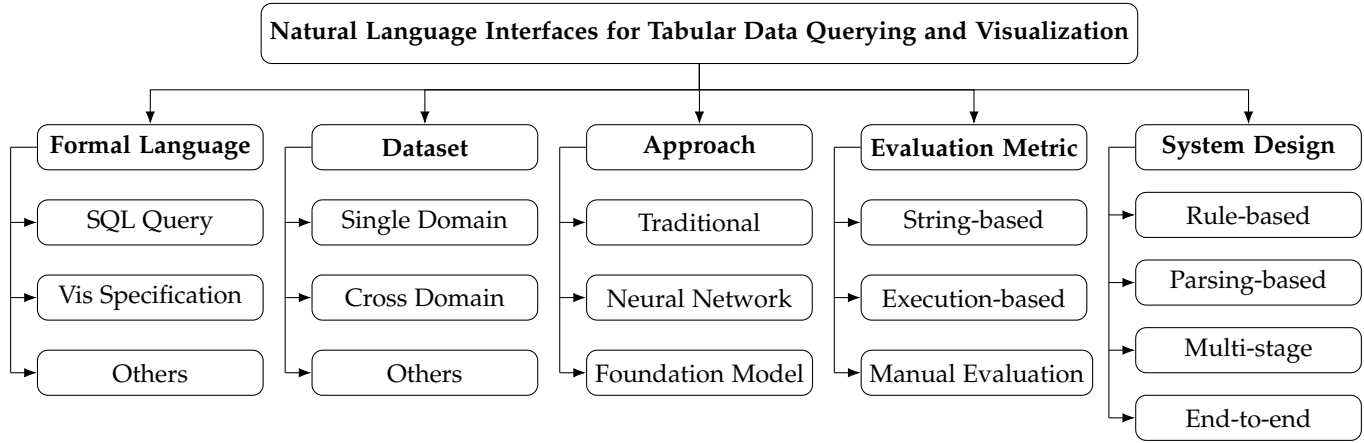

The natural language interfaces for tabular data querying and visualization encompass a variety of components, each playing a crucial role in the technology framework, as shown in Fig~\ref{fig:framework}. 


$\bullet$ \textit{Datasets.} Datasets play a vital role in training and evaluating the performance of these interfaces. Datasets can be single-turn, where a single query is posed without any prior context, or multi-turn, where a series of queries are posed in a conversational manner. There are also various types of datasets designed to evaluate different aspects of the systems, such as their ability to handle complex queries, out-of-domain queries, and more. Additionally, datasets can vary in terms of the domain they cover, such as business, healthcare, or scientific data, each posing unique challenges for natural language interfaces.

$\bullet$ \textit{Approaches.} The approaches to building natural language interfaces have evolved over time. Early approaches were rule-based, using pre-defined rules to translate natural language queries into functional representations. With the advent of neural networks, sequence-to-sequence models became popular, providing more flexibility in handling diverse queries. The rise of pre-trained language models, such as BERT~\cite{DBLP:conf/naacl/DevlinCLT19} and GPT~\cite{DBLP:conf/iclr/PatelLRCRC23}, marked a significant advancement in this field. Recently, the advent of LLMs like ChatGPT, and the exploration of prompt engineering techniques, have opened new avenues for the development of more effective natural language interfaces for data interaction. These advancements have not only improved the accuracy of these systems but have also made them more accessible to a wider range of users.

$\bullet$ \textit{Evaluation Metrics.} Evaluation metrics are used to measure the performance of these interfaces. These can be string-based, comparing the generated functional representation to a ground truth, or execution-based, comparing the result of executing the generated representation on the database to the expected result. Manual evaluation is also sometimes used to assess aspects like the system's usability. User satisfaction and usability are important evaluation criteria, as they directly impact the adoption and effectiveness of these interfaces in real-world scenarios.

$\bullet$ \textit{System Design.} System architecture is a crucial component of natural language interfaces which involves the underlying mechanisms that translate user queries into actionable outputs. The architectural paradigms, ranging from rule-based to end-to-end designs, provide varied solutions and trade-offs in terms of flexibility, interpretability, and accuracy. We also provide user-centric analysis to provide suggestions on system-selection for individual and professional users, taking into account their specific needs, technical proficiency, and the characteristics of their data environment.


Each of these components contributes to the effectiveness and usability of natural language interfaces for tabular data querying and visualization. The subsequent sections of this survey will delve into these components in more detail, discussing their role, the various methods and technologies used, and the recent advancements in each area.

\section{Datasets}
\label{section:dataset}

\subsection{Text-to-SQL Datasets}

\subsubsection{Existing Benchmarks}

Text-to-SQL datasets have evolved significantly over time, adapting to the growing complexity of the field. Early datasets are single-domain, focusing on simple, context-specific queries. As the field progressed, cross-domain datasets emerged, featuring diverse schemas and queries across multiple domains. The introduction of multi-turn conversational datasets added another layer of complexity, requiring the understanding of inter-query dependencies within a conversation. The most recent advancement is the emergence of multilingual datasets, which extend the challenge to handling queries in multiple languages. Researchers are also exploring complex scenarios such as ambiguous queries, queries requiring external knowledge, and queries involving temporal and spatial reasoning. This evolution reflects the progress and the expanding challenges in the Text-to-SQL domain. Table~\ref{tab:dataset} presents a comprehensive overview of various Text-to-SQL and Text-to-Vis datasets. \\

\noindent
\textbf{Single Domain.} The early phase of Text-to-SQL research is marked by single-domain datasets, which focus on handling queries within a specific context. \textbf{\emph{Academic}}~\cite{DBLP:journals/pvldb/LiJ14} and \textbf{\emph{Advising}}~\cite{DBLP:conf/acl/RadevKZZFRS18} are examples of early single-domain datasets. The \textbf{\emph{ATIS}} dataset~\cite{DBLP:conf/naacl/Price90,DBLP:conf/naacl/DahlBBFHPPRS94} and \textbf{\emph{GeoQuery}}~\cite{DBLP:conf/aaai/ZelleM96} are notable for their focus on flight information and U.S. geography respectively. Datasets like \textbf{\emph{Yelp}} and \textbf{\emph{IMDB}}~\cite{DBLP:journals/pacmpl/Yaghmazadeh0DD17}, \textbf{\emph{Scholar}}~\cite{DBLP:conf/acl/IyerKCKZ17}, and \textbf{\emph{Restaurants}}~\cite{DBLP:conf/emnlp/TangM00} are also developed around this time, each catering to queries pertaining to their respective domains. 
In recent years, the development of single-domain datasets has continued with the introduction of \textbf{\emph{SEDE}}~\cite{DBLP:journals/corr/abs-2106-05006} and \textbf{\emph{MIMICSQL}}~\cite{DBLP:conf/www/WangSR20}. These datasets represent the ongoing efforts to explore and address more complex and diverse queries within specific domains.\\

\noindent
\textbf{Cross Domain.} Following the single-domain datasets, the focus shifts to cross-domain datasets, which widen the scope of the Text-to-SQL task by including queries from multiple domains. A pivotal dataset marking this shift is \textbf{\emph{WikiSQL}}~\cite{DBLP:journals/corr/abs-1709-00103}. It offers a rich collection of 80,654 natural language inquiries paired with SQL queries. These pairs correspond to SQL tables extracted from a vast set of 26,521 Wikipedia tables. The dataset's uniqueness lies in its extensive coverage of tables and its capacity to challenge models to adapt to novel queries and table schemas.
Another monumental contribution to this arena is the \textbf{\emph{Spider}} dataset~\cite{DBLP:conf/emnlp/YuZYYWLMLYRZR18}. This dataset encompasses 10,181 natural language questions from 138 varied domains. Its diversity and inclusion of intricate queries make it a tougher challenge compared to its predecessors.

The Spider dataset has inspired the creation of several variants, each designed to test specific capabilities of Text-to-SQL models. For instance, \textbf{\emph{Spider-SYN}}~\cite{DBLP:conf/acl/GanCHPWXH20} tweaks the original Spider questions by substituting schema-related terms with their synonyms, elevating the schema linking challenge. \textbf{\emph{Spider-DK}}~\cite{DBLP:conf/emnlp/GanCP21} infuses domain-specific knowledge into questions, probing models' domain knowledge comprehension. Variants like \textbf{\emph{Spider-CG}}~\cite{DBLP:conf/naacl/GanCHP22} and \textbf{\emph{Spider-SSP}}~\cite{DBLP:conf/acl/ShawCPT20} focus on models' generalization abilities through diverse strategies, such as sub-sentence substitutions and compositional generalization, respectively. \textbf{\emph{Dr. Spider}}~\cite{DBLP:conf/iclr/Chang0DPZLLZJLA23} serves as a diagnostic tool, introducing variations in the original Spider dataset across multiple dimensions. Lastly, \textbf{\emph{Spider-realistic}}~\cite{DBLP:conf/naacl/DengAMPSR21} enhances task complexity by removing direct column name mentions from questions, demanding an improved robustness from models.\\

\noindent
\textbf{Multi-turn.} As the field of Text-to-SQL expanded to encompass more complex interactions, the need for datasets that could simulate multi-turn conversations became apparent. To cater to this, various datasets emphasizing context-driven Text-to-SQL interactions have been developed. \textbf{\emph{SParC}}~\cite{DBLP:conf/acl/YuZYTLLELPCJDPS19} is a prominent cross-domain dataset boasting approximately 4.3k sequences of questions, which cumulatively constitute over 12k question-SQL pairings. What's unique about SParC is that each of its question sequences evolves from an original question in Spider, with subsequent questions intricately woven in.
Similarly, the \textbf{\emph{CoSQL}} dataset~\cite{DBLP:conf/emnlp/YuZELXPLTSLJYSC19}, established under the Wizard-of-Oz framework, stands out as the first large-scale, cross-domain conversational Text-to-SQL collection. It houses nearly 3k dialogues, translating to over 30k dialogue turns and 10k associated SQL queries. Through these dialogues, the dataset replicates a scenario where annotators, posing as database users, utilize natural language to extract database responses.
Another noteworthy contribution is the \textbf{\emph{CHASE}} dataset~\cite{DBLP:conf/acl/GuoSWLFLYL20}. This dataset introduces a large-scale, context-sensitive Chinese Text-to-SQL collection, featuring 5,459 interconnected question sequences and 17,940 individual questions paired with SQL queries. Collectively, these datasets push the boundaries in the Text-to-SQL domain, emphasizing more fluid, dialogue-centric database interactions and offering diverse challenges for research exploration. \\

\begin{table*}[htbp]
\centering 
\caption{Statistics for Text-to-SQL and Text-to-Vis datasets\label{tab:dataset} sorted by ascending years and grouped by common main features. The details of each dataset include the number of NLQs (\#Query), databases (\#Database), domains (\#Domain), tables per database (\#T/DB) and the language of the queries.} 
\begin{centering}
\resizebox{0.95\textwidth}{!}{
\begin{tabular}{llllllc}
\toprule 
\textbf{Datasets} & \textbf{\#Query} & \textbf{\#Database} & \textbf{\#Domain} & \textbf{\#T/DB} & \textbf{Language} & \textbf{Main Features}  \\
\midrule 
\multicolumn{7}{c}{\textbf{Text-to-SQL Datasets}} \\
\midrule
ATIS (Hemphill et al., 1990; Dahl et al., 1994) & 5,280 & 1 & 1 & 32 & English &  \multirow{10}{*}{Single Domain}\\
GeoQuery (Zelle and Mooney, 1996)& 877 & 1 & 1 & 6 & English &  \\
Restaurants (Tang and Mooney, 2000) & 378 & 1 & 1 & 3 & English &  \\
Academic (Li and Jagadish, 2014) & 196 & 1 & 1 & 15 & English &  \\
Scholar (Iyer et al., 2017) & 817 & 1 & 1 & 7 & English &  \\
IMDB (Yaghmazadeh et al., 2017) & 131 & 1 & 1 & 16 &  English & \\
Yelp (Yaghmazadeh et al., 2017) & 128 & 1 & 1 & 7  &  English &   \\
Advising (Finegan-Dollak et al., 2018) & 3,898 & 1 & 1 & 10  &  English &  \\
MIMICSQL (Wang et al., 2020) & 10,000 & 1 & 1 & 5  &  English &   \\
SEDE (Hazoom et al., 2021) & 12,023 & 1 & 1 & 29  &  English &  \\

\midrule 
WikiSQL (Zhong et al., 2017) & 80,654 & 26,521 & - & 1 & English & \multirow{4}{*}{Cross Domain}  \\
Spider (Yu et al., 2018) & 10,181 & 200 & 138 & 5 & English & \\
Squall (Shi et al., 2020) & 11,468 & 1,679 & - & 1 & English &   \\
KaggleDBQA (Lee et al., 2021) & 272 & 8 & 8 & 2 & English &   \\
\midrule
SParC (Yu et al., 2019)&  12,726 & 200 & 138 & 5.1 & English & \multirow{3}{*}{Multi-turn} \\
CoSQL (Yu et al., 2019) &  15,598 & 200 & 138 & 5.1 & English &  \\
CHASE (Guo et al., 2021)&  17,940 &  280 & - & 4.6 & Chinese &  \\
\midrule
Spider-SYN (Gan et al., 2021) & 7,990 & 166 & - & 5 & English & \multirow{5}{*}{Robustness} \\
Spider-SSP (Shaw et al., 2021)& 3,282 & - & - & 5 & English &  \\
Spider-realistic (Deng et al., 2021)& 508 & - & - & 5 &  English &  \\
Spider-CG (Gan et al., 2022) & 45,599 & - & - & 5 & English &  \\
Dr. Spider (Chang et al., 2023)& - & 166 & - & 5 & English &  \\
\midrule
CSpider (Min et al., 2019)&  10,181  & 200 & 138 & 5 & Chinese & \multirow{6}{*}{Multilingual} \\
DuSQL (Wang et al., 2020)&  23,797 & 200 & - & 4.1 & Chinese &  \\
TableQA (Sun et al., 2020)&  64,891 &  6,029 & - & 1 & Chinese &  \\
ViText2SQL (Nguyen et al., 2020)& 9,691 & 166 & - & 5 & Vietnamese &  \\
PortugueseSpider (Archanjo Jose et al., 2021)& 9,691 & 166 & - & 5 & Portuguese &  \\
PAUQ (Bakshandaeva et al., 2022)& 9,691 & 166 & - & 5 &  Russian &  \\
\midrule
Spider-DK (Gan et al., 2021)& 535  & 10 & - & 5 & English &  \multirow{3}{*}{Knowledge Grounding } \\
knowSQL (Dou et al., 2022)& 25,888 & 200 & 3 & - & Chinese &  \\
BIRD (Li et al., 2023)& 12,751 & 95 & - & 7 & English &  \\

\midrule 
\multicolumn{7}{c}{\textbf{Text-to-Vis Datasets}} \\
\midrule 
Gao et al., 2015 & 10 & 3 & - & 1 &  English & \multirow{3}{*}{Single Domain} \\
Kumar et al., 2016 & 490 & 1 & - & -  &  English & \\
Srinivasan et al., 2021 & 893 & 3 & - & 1  &  English & \\
\midrule 
nvBench (Luo et al., 2021) & 25750 & 153 & 105 & 5 & English & Cross Domain \\
\midrule 
ChartDialogs (Shao et al., 2020) & 3284 & - & - & - & English &  \multirow{2}{*}{Multi-turn} \\
Dial-NVBench (Song et al., 2023) & 4495 & - & - & - & English &  \\
\midrule 
CNvBench (Ge et al., 2023) & 25750 & 153 & 105 & 5 & Chinese & Multilingual \\
\bottomrule
\end{tabular}}
\end{centering}
\end{table*}

\noindent
\textbf{Multilingual.} As the Text-to-SQL field expands globally, the need for multilingual datasets has become increasingly apparent. Several datasets have been developed to address this need, offering benchmarks in different languages and thereby broadening the scope of Text-to-SQL research. 
\textbf{\emph{CSpider}}~\cite{DBLP:conf/emnlp/MinSZ19}, \textbf{\emph{TableQA}}~\cite{DBLP:journals/corr/abs-2006-06434} and \textbf{\emph{DuSQL}}~\cite{DBLP:conf/emnlp/WangZWSLWZW20} extend the Text-to-SQL task to Chinese, introducing a new linguistic challenge. \textbf{\emph{ViText2SQL}}~\cite{DBLP:conf/emnlp/NguyenDN20} broadens the field further with a Vietnamese Text-to-SQL dataset, pushing models to handle the complexities of the Vietnamese language. Similarly, \textbf{\emph{PortugueseSpider}}~\cite{DBLP:conf/bracis/JoseC21} extends the task to Portuguese, requiring models to translate Portuguese queries into SQL. 
These multilingual datasets represent a significant stride towards developing Text-to-SQL systems that can cater to a global, multilingual user base, thereby democratizing access to data across linguistic boundaries.\\

\noindent
\textbf{Knowledge Grounding.} Recent advancements in Text-to-SQL research have seen a growing emphasis on knowledge-intensive benchmarks, reflecting the need for models that can handle real-world analysis scenarios. Such benchmarks, like \textbf{\emph{Spider-dk}}~\cite{DBLP:conf/emnlp/GanCP21}, extends the Spider dataset to focus more on domain knowledge, reflecting the need for models to understand and incorporate domain-specific knowledge in their translations. Another datset, \textbf{\emph{knowSQL}}~\cite{DBLP:conf/emnlp/Dou0LPWCZKL22}, prioritize knowledge grounding or commonsense reasoning, helping experts make informed decisions.
A most recent benchmark is \textbf{\emph{BIRD}}~\cite{DBLP:journals/corr/abs-2305-03111},  specifically tailored for expansive database-anchored Text-to-SQL tasks. What sets BIRD apart is its emphasis on the values within databases. It underscores novel hurdles, such as inconsistencies in database content, the imperative of bridging external knowledge with natural language queries and database content, as well as the efficiency of SQL, particularly when dealing with vast databases.
These knowledge-intensive datasets represent a significant stride towards developing Text-to-SQL systems that can handle complex, real-world scenarios, bridging the gap between academic study and practical application.

\subsection{Text-to-Vis Datasets}

\subsubsection{Existing Benchmarks}

Text-to-Vis datasets generally follow the same format as Text-to-SQL datasets with a set of tabular data and (NLQ, Vis) pairs for each database.  The progression was similarly a transition from single to cross-domain datasets though several Text-to-Vis datasets piggybacked Text-to-SQL benchmarks. \\

\noindent 
\textbf{Single Domain.} During early development stages of Text-to-Vis interfaces, datasets are generally used as a proof of concept. These datasets are each concentrated on one domain that only involves queries within a set range of context.  The dataset by Gao et al~\cite{DBLP:conf/sigdial/KumarAEJGL16} was developed by asking test subjects to pose natural utterances by looking at several human-generated visualizations with the goal to gain certain information.  The dataset by Kumar et al.~\cite{DBLP:conf/sigdial/KumarAEJGL16} focused on crime data and the queries are to gain insight to police force allocations.  \\

\noindent 
\textbf{Cross Domain.}  Srinivasan et al.~\cite{DBLP:conf/chi/SrinivasanNLDS21} collected queries across 3 datasets.  They provided thorough analyses on the classification of Text-to-Vis natural language queries. 
\textbf{\emph{nvBench}}~\cite{DBLP:journals/corr/abs-2112-12926} is the largest and most used Text-to-Vis benchmark, containing 25,750 natural language and visualization pairs from 750 tables over 105 domains.  It is synthesized from Text-to-SQL benchmark Spider~\cite{DBLP:conf/emnlp/YuZYYWLMLYRZR18} to support cross-domain Text-to-Vis task. \\

\noindent 
\textbf{Multi-turn.}  Due to the large amount of information needed to produce accurate visualizations, it is apparent that not all information may be provided in just one round of natural language query.  To tackle this issue, multi-turn datasets have been introduced to make several rounds of modifications on the output visualization.  \textbf{\emph{ChartDialogs}}~\cite{DBLP:conf/acl/ShaoN20} contains 3,284 dialogues and is curated for plotting using matplotlib.  Building on the cross-domain dataset nvBench, \textbf{\emph{Dial-NVBench}}~\cite{DBLP:journals/corr/abs-2307-16013} was created to target dialogue inputs.  The dataset contains 4,495 dialogue sessions and each is aimed to contain enough information so the system can output a suitable visualization.  \\

\noindent 
\textbf{Multilingual.} While the majority of Text-to-Vis datasets are in English, there is a growing need for multilingual datasets to support the development of natural language interfaces for diverse language communities. \textbf{\emph{CNvBench}} is a recently introduced Chinese Text-to-Vis dataset, which aims to bridge the gap between English and Chinese resources in this domain. The dataset is constructed by translating and localizing the English nvBench dataset, ensuring that the natural language queries are fluent and idiomatic in Chinese.  

\begin{tcolorbox}[colback=gray!10,colframe=black!50,title=Takeaways for Datasets]
\begin{itemize}
\item Text-to-SQL dataset evolution: Single-domain, Cross-domain (Spider), Multi-turn (SParC), Multilingual (CSpider), Knowledge-grounded (BIRD).
\item Text-to-Vis dataset evolution: Single-domain, Cross-domain (nvBench), Multi-turn (ChartDialogs), Multilingual (CNvBench)
\item Guidelines: Choose based on specific challenges and capabilities. Use Spider, nvBench for cross-domain tasks; SParC, CoSQL, ChartDialogs, Dial-NVBench for multi-turn scenarios.
\end{itemize}
\end{tcolorbox}

\section{Approaches}
\label{section:approach}

\begin{figure*}[htbp]
\centering
\includegraphics[width=0.9\textwidth]{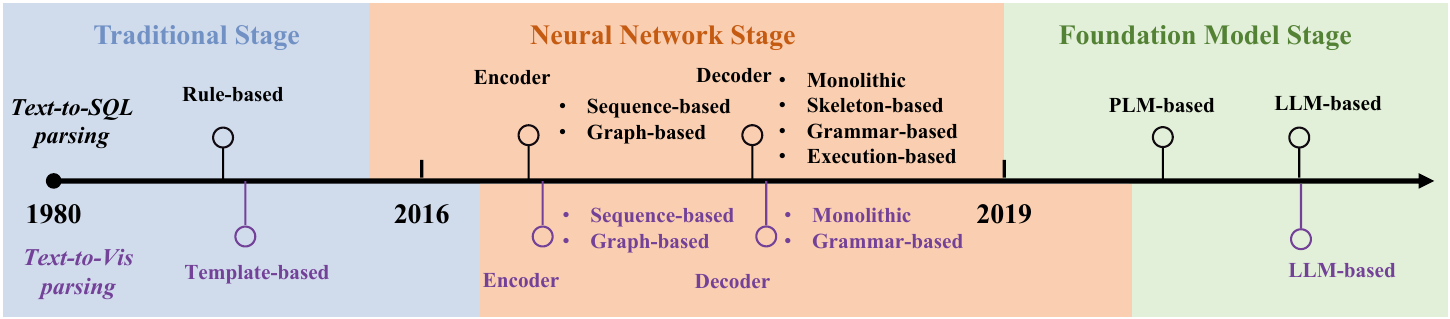}
\caption{Evolution of Text-to-SQL (upper timeline) and Text-to-Vis (lower timeline) approaches over time. The colored rectangles represent different stages of the approaches: traditional (blue), neural network (green), and foundation language model (orange). Note that the development of Text-to-Vis approaches generally occurred later than similar Text-to-SQL approaches, hence the upper and lower timelines are misaligned.}
\label{fig:timeline}
\end{figure*}

\subsection{Text-to-SQL Parsing}
The approaches to the Text-to-SQL task have evolved significantly over time, mirroring the broader developments in natural language processing, as illustrated in the timeline in Fig~\ref{fig:timeline}. Early efforts have focused on rule-based approaches, where queries are translated into SQL based on a predefined set of rules and patterns. 
The emergence of neural networks and the sequence-to-sequence paradigm mark a turning point in Text-to-SQL research. Neural network approaches, which translate a source sequence (the natural language query) into a target sequence (the SQL query), show a greater capacity to handle the intricacies of natural language and the diversity of SQL queries.
In recent years, the advent of foundation language models like BERT~\cite{DBLP:conf/naacl/DevlinCLT19} and GPT~\cite{DBLP:conf/iclr/PatelLRCRC23} has opened up new possibilities for the Text-to-SQL task. 
This evolution in approaches reflects the ongoing efforts to develop models that can accurately and efficiently translate natural language queries into SQL, handling the challenges presented by the variability of natural language and the complexity of SQL. Table~\ref{tab:approaches} provides a comparative analysis of notable approaches in the Text-to-SQL and Text-to-Vis domains.

\subsubsection{Traditional Stage}

Text-to-SQL research began with rule-based approaches, which were the primary method of handling this task for several decades. Surveys like~\cite{DBLP:conf/iui/PopescuEK03,DBLP:journals/pvldb/KimSHL20} have presented the work of this stage in more detailed ways. Early rule-based methods like \textbf{\emph{TEAM}}~\cite{DBLP:conf/anlp/Grosz83} and \textbf{\emph{CHAT-80}}~\cite{DBLP:journals/coling/WarrenP82} used intermediate logical representations, translating natural language queries into logical queries that were independent of the database schema, and then converting these logical queries into SQL. However, these methods relied heavily on hand-crafted mapping rules.

In the early 2000s, more advanced rule-based methods were developed. \textbf{\emph{PRECISE}}~\cite{DBLP:conf/coling/PopescuAEKY04} utilized an off-the-shelf natural language parser to translate queries, but its coverage was limited due to the assumption of a one-to-one correspondence between words in the query and database elements. To address this, methods like \textbf{\emph{NaLIR}}~\cite{DBLP:journals/pvldb/LiJ14}, \textbf{\emph{ATHENA}}~\cite{DBLP:journals/pvldb/SahaFSMMO16}, and \textbf{\emph{SQLizer}}~\cite{DBLP:journals/pacmpl/Yaghmazadeh0DD17} adopted a ranking-based approach, finding multiple candidate mappings and ranking them based on a score. NaLIR further improved performance by involving user interaction, while ATHENA leveraged a domain-specific ontology for richer semantic information. SQLizer used an iterative process to refine the logical form of the query.
\textbf{\emph{Templar}}~\cite{DBLP:conf/icde/BaikJ019} offered an optimization technique for mapping and joint path generation using a query log. Despite their significant improvements, these methods still relied on manually-defined rules, which limited their ability to handle many variations in natural language.

\subsubsection{Neural Network Stage}

The advent of neural networks and the sequence-to-sequence (Seq2Seq) paradigm has marked a turning point in the field of Text-to-SQL. Originally for machine translation, Seq2Seq models can learn intricate data mappings, accommodating diverse queries and complex SQL structures. Such a model typically uses an encoder to process the natural language query and a decoder to generate the corresponding SQL query. For a deeper dive into neural network-based approaches, readers are encouraged to consult prior surveys like~\cite{DBLP:journals/vldb/KatsogiannisMeimarakisK23,DBLP:conf/sigmod/Katsogiannis-Meimarakis21,DBLP:conf/coling/Deng0022}.\\

\noindent
\textbf{Encoder.} Encoders in the Text-to-SQL context determine how the natural language query and the database schema are jointly transformed into a continuous representation that the model can work with. They can broadly be classified into two categories: sequence-based encoders and graph-based encoders.

$\bullet$ \textit{Sequence-based Encoder.} Sequence-based encoders form the foundation of many Text-to-SQL systems. They are often based on Recurrent Neural Networks (RNNs), Long Short-Term Memory (LSTM) networks, Gated Recurrent Units (GRUs), or Transformer architectures.

Bi-directional Long Short-Term Memory (bi-LSTM) based models have been widely used in early Text-to-SQL systems due to their capability to capture dependencies in both directions of a sequence. Notable work includes \textbf{\emph{TypeSQL}}~\cite{DBLP:conf/naacl/YuLZZR18}, which assigns a type to each word in the question, with a word being an entity from the knowledge graph, a column, or a number. The model then concatenates word embeddings and the corresponding type embeddings as input to the bi-LSTM, which helps it better encode keywords in questions. \textbf{\emph{Seq2SQL}}~\cite{DBLP:journals/corr/abs-1709-00103}, \textbf{\emph{SQLNet}}~\cite{DBLP:journals/corr/abs-1711-04436}, and \textbf{\emph{IncSQL}}~\cite{DBLP:journals/corr/abs-1809-05054} employ a bi-LSTM to produce a hidden state representation for each word in the natural language query. For column headers, a bi-LSTM is also used for each column name, with the final hidden state used as the initial representation for the column. For example, \textbf{\emph{EditSQL}}~\cite{DBLP:conf/emnlp/ZhangYESXLSXSR19} also utilizes two separate Bi-LSTMs for encoding the natural language questions and the table schema, and then applies a dot-product attention layer to integrate the two encodings.

With the advent of the Transformer architecture, self-attention models have gained popularity in the Text-to-SQL task. The original self-attention mechanism is the building block of the Transformer structure, and models like those developed by He et al.~\cite{DBLP:journals/corr/abs-1908-08113}, Hwang et al.~\cite{DBLP:journals/corr/abs-1902-01069}, and Xie et al.~\cite{DBLP:conf/emnlp/XieW0ZSYWZYWZWL22}, have incorporated this mechanism. These models leverage the Transformer's ability to capture dependencies regardless of their distance in the sequence, which is especially useful for handling complex, non-local dependencies often present in the Text-to-SQL task.

$\bullet$ \textit{Graph-based Encoder.} Graphs are an effective way to capture complex structures, making them particularly suitable for encoding database (DB) schemas, which are rich in structural information. 
Bogin et al.~\cite{DBLP:conf/acl/BoginBG19} were pioneers in using graph representations for DB schemas. They employed nodes for tables and columns and edges to depict table-column relationships, such as table compositions, and primary and foreign key constraints. These graph structures were then encoded using graph neural networks (GNNs).
In a follow-up study,  Bogin et al. introduced \textbf{\emph{Global-GNN}}~\cite{DBLP:conf/emnlp/BoginGB19}, emphasizing global reasoning to encode the schema, integrating question token representations between question terms and schema entities.
\textbf{\emph{RAT-SQL}}~\cite{DBLP:conf/acl/WangSLPR20} combined global reasoning, structured reasoning, and relation-aware self-attention for schema entities and question terms.

Graphs have also been employed to concurrently encode questions and DB schemas. Cao et al. put forward the \textbf{\emph{LGESQL}}~\cite{DBLP:conf/acl/CaoC0ZZ020} model to unearth multi-hop relational attributes and significant meta-paths. 
\textbf{\emph{S$^2$SQL}}~\cite{DBLP:conf/acl/HuiGWQLLSL22} explored question token syntax's role in Text-to-SQL encoders and introduced a versatile and resilient injection technique. 
To strengthen the graph method's generalization for unfamiliar domains, \textbf{\emph{SADGA}}~\cite{DBLP:conf/nips/CaiYXH21} crafted both question and schema graphs based on the dependency structure of natural language queries and schema layout, respectively. 
\textbf{\emph{ShawdowGNN}}~\cite{DBLP:conf/naacl/ChenCZCXZY21} countered domain information influenced by disregarding table or column names and employing abstract schemas for delexicalized representations. Lastly, Hui et al. 2021~\cite{DBLP:conf/aaai/HuiGRLLSHSZZ21} designed a dynamic graph framework to capture interactions among utterances, tokens, and database schemas, leveraging both context-independent and dependent parsing.\\

\noindent
\textbf{Decoder.} The decoder is a crucial component of the sequence-to-sequence paradigm, responsible for generating the SQL query from the encoded representation of the natural language query and database schema. 
Broadly, these decoders can be classified into four categories: monolithic decoders, skeleton-based decoders, grammar-based decoders, and execution-guided decoders.

$\bullet$ \textit{Monolithic Decoder.} The Monolithic decoder, influenced by advancements in machine translation, primarily utilizes RNNs for the sequential generation of SQL commands. Early implementations of this method relied on RNNs to compute the probability of each SQL token, considering both the prior context and previously generated tokens~\cite{DBLP:conf/acl/IyerKCKZ17}. The context from the input is encoded, often using mechanisms like soft-attention, which emphasizes the most pertinent input components for each token generation. As for representing previously generated tokens, a common method is to use hidden states from the prior decoder step. 

\begin{table*}[htbp]
\caption{Approaches for natural language interfaces for database querying and visualization. The approaches are categorized into three stages: Traditional, Neural Network, and Foundation Language Model. The "WikiSQL" and "Spider" columns report the execution accuracy (EX) and exact match (EM) on the dev sets of the respective datasets, which are widely used benchmarks for Text-to-SQL tasks. WikiSQL focuses on single-table queries, while Spider covers more complex multi-table queries. The "NVBench" column reports the overall accuracy (exact match) on the NVBench dataset, a benchmark for Text-to-Vis tasks. "-" indicates that the metric is not reported or the approach is not evaluated on that dataset.}\label{tab:approaches}
\begin{adjustbox}{width=\textwidth,keepaspectratio}
\begin{tabular}{lcccclc}
\toprule
\multirow{2}{*}{\textbf{Approach}}
& \multirow{2}{*}{\textbf{Task}} & \textbf{WikiSQL} & \textbf{Spider} & \textbf{NVBench} &  \multirow{2}{*}{\textbf{Description}} & \multirow{2}{*}{\textbf{Key Features}} \\
& & \textbf{EX(\%)} & \textbf{EM(\%)} & \textbf{Acc.(\%)} & & \\
\midrule
\multicolumn{7}{c}{\textbf{Traditional Stage}} \\
\midrule

CHAT-80 (Warren and Pereira, 1982) & Query & - & - & - & English to logic using extraposition grammars & \multirow{6}{*}{Rule-based} \\
TEAM (Grosz, 1983) & Query & - & - & - & Adapts to DBs using intermediate logic & \\
PRECISE (Popescu et al., 2004) & Query & - & - & - & Semantic-enhanced statistical parser & \\
NaLIR (Li and Jagadish, 2014) & Query & - & - & - & Guided user interaction for complex queries & \\
ATHENA (Saha et al., 2016) & Query & - & - & - & Domain ontology for query translation & \\
Templar (Baik et al., 2019) & Query & - & - & - & Optimized keyword mapping and join path & \\
\midrule
SQLizer (Yaghmazadeh et al., 2017) & Query & - & - & - & Program synthesis for logical form refinement & \multirow{4}{*}{Template-based} \\
DataTone (Gao et al., 2015) & Visual & - & - & - & Interactive widgets for NL ambiguity & \\
ADVISor (Liu et al., 2021) & Visual & - & - & - & NL-based annotated visualizations & \\
NL4DV (Narechania et al., 2021) & Visual & - & - & - & NL to analytic specs toolkit & \\

\midrule
\multicolumn{7}{c}{\textbf{Neural Network Stage}} \\
\midrule

SQLNet (Xu et al., 2017) & Query & 69.8 & - & - & Seq-to-set, column attention & \multirow{8}{*}{Seq-based Enc.}\\
TypeSQL (Yu et al., 2018) & Query & 85.5 & - & - & Type-aware entity and value understanding & \\
IncSQL (Shi et al., 2018) & Query & 87.2 & - & - & Incremental seq-to-action SQL construction & \\

EditSQL (Zhang et al., 2019) & Query & - & 57.6 & - & Query generation via editing & \\

Data2Vis (Dibia and Demiralp, 2019) & Visual & - & - & - & End-to-end neural vis generation & \\
Seq2Vis (Luo et al., 2021) & Visual & - & - & 1.95 & Pre-trained word embeddings for input & \\
ncNet (Luo et al., 2022) & Visual & - & - & 25.78 & Vis-aware Transformer optimizations &\\
MMCoVisNet (Song et al., 2023) & Visual & - & - & - & Multi-modal dialogue context understanding & \\
\midrule
GNN (Bogin et al., 2019) & Query & - & 40.7 & - & Graph neural network for DB schema encoding & \multirow{9}{*}{Graph-based Enc.} \\
Global-GNN (Bogin et al., 2019) & Query & - & 52.1 & - & Global reasoning for query structure & \\
RAT-SQL (Wang et al., 2020) & Query & - & 69.7 & - & Relation-aware schema encoding and linking & \\
LGESQL (Cao et al., 2021) & Query & - & 75.1 & - & Line graph encoding for message passing & \\

SADGA (Cai et al., 2021) & Query & - & 73.1 & - & Dual graph aggregation for question-schema & \\
ShadowGNN (Chen et al., 2021) & Query & - & 72.3 & - & Graph projection for delexicalized representations & \\
Hui et al., 2021 & Query & - & - & - & Dynamic graph for contextual relations & \\

$S^2SQL$ (Hui et al., 2022) & Query & - & 76.4 & - & Syntactic info for question-schema encoding & \\

RGVisNet (Song et al., 2023) & Visual & - & - & 44.9 & Retrieval-generation for data vis & \\

\midrule

COARSE2FINE (Dong and Lapata, 2018) & Query & 79.6 & - & - & Coarse-to-fine semantic parsing & \multirow{4}{*}{Skeleton-based Dec.} \\

IE-SQL (Ma et al., 2020) & Query & 92.6 & - & - & Extraction-linking for slot filling & \\
HydraNet (Lyu et al., 2020) & Query & 92.4 & - & - & Separate column ranking and decoding & \\
RYANSQL (Choi et al., 2021) & Query & - & 66.6 & - & Sketch-based slot filling with SPCs & \\

\midrule
Seq2Tree (Dong et al., 2016) & Query & - & - & - & Encoded repr. for seq-to-tree generation & \multirow{8}{*}{Grammar-based Dec.} \\
Seq2AST (Yin and Neubig, 2017) & Query & - & - & - & Target syntax as prior knowledge & \\
SyntaxSQLNet (Yu et al., 2018) & Query & - & 24.8 & - & SQL-specific syntax tree decoder & \\

IRNet (Guo et al., 2019) & Query & - & 61.9 & - & Intermediate repr. bridging NL and SQL &\\

SmBoP (Rubin and Berant, 2021) & Query & - & 69.5 & - & Semi-autoregressive bottom-up parsing & \\
NatSQL (Gan et al., 2021) & Query & - & 73.7 & - & Simplified IR for complex SQL & \\
PICARD (Scholak et al., 2021) & Query & - & 75.5 & - & Constrained autoregressive decoding & \\
UniSAr (Dou et al., 2022) & Query & 91.7 & 70.0 & - & Unified structure-aware autoregressive & \\

\midrule
Seq2SQL (Zhong et al., 2017) & Query & 60.8 & - & - & Execution-guided SQL generation & \multirow{3}{*}{Execution-based Dec.} \\

Wang et al., 2018 & Query & 78.5 & - & - & Runtime error correction, ensemble & \\
Suhr et al., 2020 & Query & - & 65.8 & - & Cross-domain semantic parsing evaluation & \\

\midrule
\multicolumn{7}{c}{\textbf{Foundation Language Model Stage}} \\
\midrule

SQLOVA (Hwang et al., 2019) & Query & 90.2 & - & - & BERT with schema-aware contextualization & \multirow{9}{*}{PLM-based}\\
X-SQL (He et al., 2019) & Query & 92.3 & - & - & PLM-augmented schema representation & \\
TaBERT (Yin et al., 2020) & Query & - & 65.2 & - & Joint text-table pre-training & \\
Bridge (Lin et al., 2020) & Query & - & 71.1 & - & BERT-contextualized question and schema & \\
GraPPa (Yu et al., 2021) & Query & 82.2 & 73.4 & - & Grammar-guided pre-training & \\
GAP (Shi et al., 2021) & Query & - & 71.8 & - & Generation-augmented pre-training & \\

UnifiedSKG (Xie et al., 2022) & Query & 85.96 & 71.76 & - & T5 benchmarking for SKG tasks & \\

Graphix-T5 (Li et al., 2023) & Query & - & 77.1 & - & Graph-aware T5 for complex reasoning & \\
RESDSQL (Li et al., 2023) & Query & - & 80.5 & - & Ranking-enhanced schema linking & \\

\midrule

C3 (Dong et al., 2023) & Query & - & - & - & Zero-shot ChatGPT with calibration & \multirow{14}{*}{LLM-based} \\
ZERoNL2SQL (Gu et al., 2023) & Query & - & - & - & PLM sketch, LLM reasoning & \\
DIN-SQL (Li et al., 2023) & Query & - & 60.1 & - & Task decomposition for few-shot LLM & \\
Liu and Tan, 2023 & Query & - & - & - & Chain-of-Thought LLM prompting & \\
SC-Prompt (Gu et al., 2023) & Query & - & 76.9 & - & Structure-content decoupling, hybrid prompting & \\
Nan et al., 2023 & Query & - & - & - & In-context learning with demo selection & \\
Tai et al., 2023 & Query & - & - & - & Simplified CoT for Text-to-SQL & \\
SQL-PaLM (Sun et al., 2023) & Query & - & - & - & Self-consistency prompting with PaLM-2 & \\
Guo et al., 2023 & Query & - & - & - & Retrieval-augmented iterative refinement & \\
DAIL-SQL (Gao et al., 2023 ) & Query & - & - & - & Systematic LLM prompt engineering & \\
PET-SQL (Li et al., 2024 ) & Query & - & - & - & Schema-aware cross-model consistency & \\
NL2INTERFACE (Chen et al., 2022) & Visual & - & - & - & NL-based interactive vis interface & \\
Chat2VIS (Maddigan and Susnjak, 2023) & Visual & - & - & - & Prompt-engineered LLMs for vis code & \\
Prompt4Vis (Li et al., 2024) & Visual & - & - & 52.69 & In-context learning for text-to-vis & \\
\bottomrule
\end{tabular}%
\end{adjustbox}
\end{table*}

$\bullet$ \textit{Skeleton-based Decoder.} Skeleton-based decoders tackle the Text-to-SQL problem by first generating a template or skeleton of the SQL query, which is then populated with specific details from the input. This approach can help manage the complexity of SQL queries by breaking down the generation process into more manageable steps.
For example, \textbf{\emph{SQLNet}}~\cite{DBLP:journals/corr/abs-1711-04436} introduced the approach that focuses on filling in the slots in a SQL sketch, aligning with the SQL grammar, rather than predicting both the output grammar and the content. This approach captures the dependency of the predictions, where the prediction of one slot is conditioned only on the slots it depends on. \textbf{\emph{HydraNet}}~\cite{DBLP:journals/corr/abs-2008-04759} uses a multi-headed selection network for simultaneous generation of different parts of the SQL query. 
\textbf{\emph{IE-SQL}} and \textbf{\emph{TypeSQL}}~\cite{DBLP:conf/naacl/YuLZZR18} also use a slot-filling approach, where a pre-defined SQL template is filled in based on the input. 
\textbf{\emph{COARSE2FINE}}~\cite{DBLP:conf/acl/LapataD18} adopts a two-step coarse-to-fine generation process, where an initial rough sketch is generated and subsequently refined with low-level details conditioned on the question and the sketch.
\textbf{\emph{RYANSQL}}~\cite{DBLP:journals/coling/ChoiSKS21} takes a recursive approach to yield SELECT statements and employs a sketch-based slot filling for each of the SELECT statements. This approach effectively handles complex SQL queries with nested structures.


$\bullet$ \textit{Grammar-based Decoder.} Grammar-based decoders generate the SQL query directly from the encoded representation of the input, often utilizing SQL grammar rules, intermediate representations, or incorporating constraints in the decoding process to ensure the generation of valid SQL queries.

Decoders utilizing rules aim to reduce the chances of generating out-of-place tokens or syntactically incorrect queries. By generating a sequence of grammar rules instead of simple tokens, these models ensure the syntactical correctness of the generated SQL queries. For example, \textbf{\emph{Seq2Tree}}~\cite{DBLP:conf/acl/DongL16} employs a top-down decoding strategy, generating logical forms that respect the hierarchical structure of SQL syntax. 
\textbf{\emph{Seq2AST}}~\cite{DBLP:conf/acl/YinN17} takes this idea further by using an abstract syntax tree (AST) for decoding. 
\textbf{\emph{SyntaxSQLNet}}~\cite{DBLP:conf/emnlp/YuYYZWLR18} adapts this approach to SQL-specific syntax. It employs a tree-based decoder that recursively calls modules to predict different SQL components, providing a structured approach to SQL query generation.
\textbf{\emph{SmBoP}}~\cite{DBLP:conf/acl-spnlp/RubinB21} stands out for its bottom-up decoding mechanism. Given a set of trees, it scores and selects trees based on SQL grammar, ensuring that the generated queries are both syntactically valid and semantically aligned with the input.
\textbf{\emph{Bridge}}~\cite{DBLP:journals/corr/abs-2012-12627} uses an LSTM-based pointer-generator with multi-head attention and a copy mechanism as the decoder. This model is capable of generating a token from the vocabulary, copying a token from the question, or copying a schema component from the database schema at each decoding step, providing a flexible approach to SQL query generation.

Some other decoders generate an intermediate representation (IR) of the SQL query first, simplifying the SQL generation task by breaking it down into more manageable steps. 
Typical models include \textbf{\emph{IncSQL}}~\cite{DBLP:journals/corr/abs-1809-05054} which defines distinct actions for different SQL components and lets the decoder predict these actions instead of directly generating SQL queries, effectively simplifying the generation task.
\textbf{\emph{IRNet}}~\cite{DBLP:conf/acl/GuoZGXLLZ19} introduces SemQL, an intermediate representation for SQL queries. SemQL can cover a wide range of SQL queries, making it a versatile tool for SQL query generation.
\textbf{\emph{NatSQL}}~\cite{DBLP:conf/emnlp/GanCXPWDZ21} builds on the idea of SemQL by removing set operators, streamlining the IR and making it easier to handle.

There are also constrained decoding-based decoders, which incorporate constraints into the decoding process to guide the SQL query generation. 
Models like \textbf{\emph{PICARD}}~\cite{DBLP:conf/emnlp/ScholakSB21} and \textbf{\emph{UniSAr}}~\cite{DBLP:journals/corr/abs-2203-07781} use reinforcement learning and rule-based systems, respectively, to incorporate constraints into the decoding process. These constraints guide the model towards generating valid SQL queries, contributing to the accuracy and reliability of these models.

$\bullet$ \textit{Execution-based Decoder.} Execution-based decoders offer a unique approach to the Text-to-SQL task, utilizing an off-the-shelf SQL executor such as SQLite to verify the validity and correctness of the generated SQL queries during the decoding process. This methodology ensures both syntactic and semantic accuracy of the produced SQL queries.
Wang et al.~\cite{DBLP:journals/corr/abs-1807-03100} leverages a SQL executor to check the partially generated SQL queries during the decoding process. Queries that raise errors are discarded, and the model continues to refine the generation until a valid SQL query is produced. 
Suhr et al.~\cite{DBLP:conf/acl/SuhrCSL20} follow a similar approach, but they avoid altering the decoder's structure. Instead, they examine the executability of each candidate SQL query. Only the queries that can be successfully executed are considered valid, which helps in maintaining the grammatical correctness of the generated SQL queries.
In another approach, \textbf{\emph{SQLova}}~\cite{DBLP:journals/corr/abs-1902-01069} incorporate an execution-guided decoding mechanism that filters out non-executable partial SQL queries from the output candidates. This methodology ensures the generation of SQL queries that are not only syntactically correct but also executable.

\subsubsection{Foundation Language Model Stage}

The recent upsurge in the performance of NLP tasks is significantly attributed to the advancement of foundation language models (FMs) such as BERT, T5, and GPT. These models, trained on large corpora, capture rich semantic and syntactic features of languages and have been successful across a variety of tasks. 

We categorize the FM-based approaches in Text-to-SQL into two categories based on the language models they incorporate: Pretrained Language Models(PLMs) and Large Language Models(LLMs).
PLMs, representing the earlier evolution like BERT and initial GPT versions, capture detailed linguistic nuances through extensive training. They are often refined for specific tasks via methods like fine-tuning.
LLMs represent an advancement, characterized by their vast scale. By amplifying model parameters or training data, these models exhibit enhanced "emergent abilities"~\cite{DBLP:journals/tmlr/WeiTBRZBYBZMCHVLDF22}. A prime example is ChatGPT, an adaptation of the GPT architecture that excels in dialogue interactions. LLM-based Text-to-SQL methods leverage prompts, utilizing in-context learning~\cite{dong2023survey} and chain-of-thought~\cite{DBLP:conf/nips/Wei0SBIXCLZ22} reasoning to produce apt SQL queries.\\

\noindent
\textbf{PLM-based.} Early PLM-based approaches directly utilize and fine-tune pre-trained language models, refining them specifically for the Text-to-SQL task. These models can be broadly categorized into encoder-only language models and encoder-decoder language models.

$\bullet$ \textit{Encoder-only Language Models.} Models like BERT and RoBERTa~\cite{DBLP:journals/corr/abs-1907-11692} serve as foundational encoder-only PLMs in various Text-to-SQL models, transforming input sequences into context-sensitive numerical representations.
\textbf{\emph{IRNet}}~\cite{DBLP:conf/acl/GuoZGXLLZ19}, for example, harnesses BERT to craft a specialized input sequence.
\textbf{\emph{BRIDGE}}~\cite{DBLP:journals/corr/abs-2012-12627} fuses BERT's prowess with schema-consistency guided decoding in a seq-to-seq architecture, enhancing the schema linking ability.
\textbf{\emph{HydraNet}}~\cite{DBLP:journals/corr/abs-2008-04759} and \textbf{\emph{SQLova}}~\cite{DBLP:journals/corr/abs-1902-01069} process questions and columns separately, predicting for each column individually with BERT, notably excelling on the WikiSQL benchmark.
\textbf{\emph{X-SQL}}~\cite{DBLP:journals/corr/abs-1908-08113} makes a novel modification to BERT by replacing segment embeddings with column type embeddings. This model also encodes additional feature vectors for matching question tokens with table cells and column names, and concatenates them with BERT embeddings of questions and database schemas.


$\bullet$ \textit{Encoder-decoder Language Models.} Unlike encoder-only models, encoder-decoder models like T5~\cite{DBLP:journals/jmlr/RaffelSRLNMZLL20} and BART~\cite{DBLP:conf/acl/LewisLGGMLSZ20} are end-to-end models designed for seq-to-seq tasks. These models take a sequence of textual input and generate a sequence of textual output. They have been adapted and fine-tuned for the Text-to-SQL task, resulting in innovative and effective models.
\textbf{\emph{UnifiedSKG}}~\cite{DBLP:conf/emnlp/XieW0ZSYWZYWZWL22}, for example, fine-tunes T5 on Text-to-SQL task with PICARD~\cite{DBLP:conf/emnlp/ScholakSB21} decoding. By combining the advantages of T5's powerful language understanding capabilities with the benefits of a sketch-based approach, it captures both the structural aspects of SQL and the semantic nuances of natural language questions. 
\textbf{\emph{Graphix-T5}}~\cite{DBLP:conf/aaai/LiHCQ0HHDSL23} leverages the robust contextual encoding intrinsic of T5 to enhance domain generalization by modeling relational structures. Its GRAPHIX layer encodes both semantic and structural insights, marking a pioneering step in infusing graphs into Text-to-SQL translation.
\textbf{\emph{RESDSQL}}~\cite{DBLP:conf/aaai/Li00023} also taps into the T5 model to craft the SQL query, utilizing a fusion of the question and schema sequences, where various T5 variants are adapted to generate skeletons derived from questions.

$\bullet$ \textit{Additional Pretraining.} Other than finetuning from general pretrained language models, there are some approaches involving additional pretraining of language models with Text-to-SQL data. Rather than directly employing off-the-shelf PLMs, these methods construct a new model using architectures like BERT or BART, and train these models using Text-to-SQL data (tabular data and text-to-SQL pairs) with specially designed objectives that are related to SQL generation. 

For instance, \textbf{\emph{TaBERT}}~\cite{DBLP:conf/acl/YinNYR20} enhances BERT by training on tabular data, focusing on predicting concealed column names and restoring cell values. This equips the model with insights into database tables' structure and content, which is crucial for accurate SQL query generation.
\textbf{\emph{Grappa}}~\cite{DBLP:conf/iclr/0009WLWTYRSX21} finetunes BERT by generating question-SQL pairs over tables. The training targets objectives like masked language modeling (MLM), column prediction, and SQL operation prediction, honing the model's ability to produce SQL queries aligned with the natural language intent.
\textbf{\emph{GAP}}~\cite{DBLP:conf/aaai/ShiNWZLWSX21} follows a parallel strategy, pretraining BART on combined Text-to-SQL and tabular datasets. The training focuses on objectives like MLM, predicting columns, restoring columns, and crafting SQL. Integrating these goals, GAP ensures that the model comprehends subtle differences in the database tables and the posed questions, improving the precision of generated SQL queries.\\

\noindent
\textbf{LLM-based.}  LLM-based methods mark the latest trend in Text-to-SQL, combining the power of large language models with the art of prompt engineering. These approaches use carefully designed prompts to steer the models toward generating accurate SQL queries, with two main categories: zero-shot prompting and few-shot prompting. 

$\bullet$ \textit{Zero-shot Prompting.} In zero-shot prompting, the LLM receives a specific prompt without any additional training examples, banking on the extensive knowledge it has gained during the pre-training phase. Rajkumar et al.~\cite{DBLP:journals/corr/abs-2204-00498} first embarked on an empirical exploration of zero-shot Text-to-SQL capabilities on Codex~\cite{DBLP:journals/corr/abs-2107-03374}. After ChatGPT came out, Liu et al.~\cite{DBLP:journals/corr/abs-2303-13547} conducted an extensive evaluation of zero-shot Text-to-SQL ability on it across an array of benchmark datasets. Building on this, the method \textbf{\emph{C3}} ~\cite{DBLP:journals/corr/abs-2307-07306} based on ChatGPT emerged as a leading zero-shot Text-to-SQL solution on the Spider Challenge. The essence of C3 lies in its three foundational prompting components: Clear Prompting, Calibration with Hints, and Consistent Output.
\textbf{\emph{ZERoNL2SQL}}~\cite{DBLP:journals/corr/abs-2306-08891} has merged the strengths of both PLMs and LLMs to foster zero-shot Text-to-SQL capabilities. The approach leverages PLMs for the generation of an SQL sketch through schema alignment and subsequently employs LLMs to infuse the missing details via intricate reasoning. A distinctive feature of their method is the predicate calibration, designed to align the generated SQL queries closely with specific database instances.

$\bullet$ \textit{Few-shot Prompting.} Few-shot prompting in Text-to-SQL presents a fascinating landscape where models are guided to achieve complex tasks with minimal examples. The strategies of in-context learning (ICL) and chain-of-thought (CoT) reasoning play pivotal roles in these approaches, enabling models to extract knowledge from a handful of demonstrations and reason through intricate SQL generation processes.

A notable work in this area is \textbf{\emph{DIN-SQL}}~\cite{DBLP:journals/corr/abs-2304-11015}, which showcases how breaking down SQL query generation into constituent problems can significantly improve the performance of LLMs. This is achieved through a four-module strategy: schema linking, query classification and decomposition, SQL generation, and a novel self-correction mechanism.
Similarly, Liu et al.~\cite{DBLP:journals/corr/abs-2304-11556} brought forth the \textit{Divide-and-Prompt} paradigm which decomposes the primary task into simpler sub-tasks, tackling each through a CoT approach, thereby enhancing the reasoning abilities of LLMs for the Text-to-SQL task. Gu et al.~\cite{DBLP:journals/pacmmod/GuF00JM023} presented a unique Divide-and-Conquer framework which steers LLMs to generate structured SQL queries and subsequently populates them with concrete values, ensuring both validity and accuracy.

In a comprehensive study, Nan et al.~\cite{DBLP:journals/corr/abs-2305-12586} explored various prompt design strategies to enhance Text-to-SQL models. The research probes into different demonstration selection methods and optimal instruction formats, revealing that a balance between diversity and similarity in demonstration selection combined with database-related knowledge augmentations can lead to superior outcomes. Tai et al.~\cite{DBLP:journals/corr/abs-2305-14215} proposed a systematic investigation of enhancing LLM's reasoning abilities for text-to-SQL parsing through various chain-of-thought style promptings. The research found that avoiding excessive detail in reasoning steps and improving multi-step reasoning can lead to superior results.

More recently, \textbf{\emph{SQL-PaLM}}~\cite{DBLP:journals/corr/abs-2306-00739}, an LLM-based approach grounded in PaLM-2, is proposed employing an execution-based self-consistency prompting approach tailored for Text-to-SQL. Guo et al.~\cite{DBLP:journals/corr/abs-2307-05074} propose a retrieval-augmented prompting method that integrates sample-aware demonstrations and a dynamic revision chain. This approach aims to generate executable and accurate SQLs by iteratively adapting feedback from previously generated SQL, ensuring accuracy without human intervention.

\subsection{Text-to-Vis Parsing}

Currently, there are several models specifically handling the Text-to-Vis problem. They typically accept a natural language query and tabular data, producing a self-defined visual language query (VQL), a SQL-like pseudo syntax for combining database querying with visualization directives, which is then hard-coded to visual specification code.  Similar to Text-to-SQL, Text-to-Vis parsing approaches have transitioned through three evolutionary stages: traditional, neural network, and foundation language model, as illustrated in Fig.~\ref{fig:timeline}.

\subsubsection{Traditional Stage}
During this stage, the main focus was to improve accuracy by using different parsing methods, keywords, and grammar rules.  Between 2015 and 2020, the works mostly explored the effects of different semantic parsing techniques. Notable works include \textbf{\emph{DataTone}}~\cite{DBLP:conf/uist/GaoDALK15}, \textbf{\emph{Eviza}}~\cite{DBLP:conf/uist/SetlurBTGC16}, \textbf{\emph{Evizeon}}~\cite{DBLP:journals/tvcg/HoqueSTD18}, \textbf{\emph{VisFlow}}~\cite{DBLP:journals/tvcg/YuS17}, \textbf{\emph{FlowSense}}~\cite{DBLP:journals/tvcg/YuS20}, \textbf{\emph{Orko}}~\cite{DBLP:journals/tvcg/SrinivasanS18}, \textbf{\emph{Valletto}}~\cite{DBLP:conf/chi/KasselR18}, and \textbf{\emph{InChorus}}~\cite{DBLP:conf/chi/SrinivasanLRDH20}.  The survey by Shen et al.~\cite{DBLP:journals/tvcg/ShenSLYHZTW23} gave a thorough walk-through of the different methods.  Stemming from the method in DataTone, several works in 2020 and 2021 have deployed a more structured VQL template.  The VQLs for each system are defined slightly differently but they generally follow the SQL style and include additional visualization attributes. \textbf{\emph{ADVISor}} ~\cite{DBLP:conf/apvis/0004HJY21} has developed an automatic pipeline to generate visualization with annotations.  The input is a set of table headers and an NLQ and the output is a set of aggregations in a SQL-like format.  \textbf{\emph{NL4DV}} ~\cite{DBLP:journals/tvcg/NarechaniaSS21} has provided a python package that takes an NLQ and the associated tabular dataset as input and outputs visualization recommendations in the form of a JSON object that can help users generate visualizations.

\subsubsection{Neural Network Stage}
The emergence of deep neural networks, especially attention mechanisms, brings a shift towards encoder-decoder-based models. As discussed earlier, the template approach can be easily converted to a neural network model. In some models, visualization specifications are directly produced, bypassing the intermediate VQL sequence step. This section delves into various models leveraging the encoder-decoder architecture.\\

\noindent \textbf{Encoder.} Sequence-based encoders like LSTMs and transformers excel at managing sequential data's long-term dependencies, while graph-based encoders grasp non-linear relationships, comprehensively depicting the input. Their capability to represent complex data structures establishes their significance in crafting efficient Text-to-Vis systems.

$\bullet$ \textit{Sequence-based Encoder.} Sequence-based encoders like LSTM, attention mechanisms, and transformers have become essential to Text-to-Vis. While LSTMs are great at managing sequential long-term dependencies, they are restricted in modeling complex interactions between distant words. This limitation is addressed by the attention mechanism and is further enhanced by the Transformer architecture. 

\textbf{\emph{Seq2Vis}}~\cite{DBLP:conf/sigmod/Luo00CLQ21}, evolving from \textbf{\emph{Data2Vis}}~\cite{DBLP:journals/cga/DibiaD19}, employs a seq2seq model, enhancing it with pretrained global word embeddings for richer input understanding. Combined with LSTM encoders, attention, and LSTM decoders, Seq2Vis adeptly translates natural language queries into visualizations. Similarly, \textbf{\emph{MMCoVisNet}}~\cite{DBLP:journals/corr/abs-2307-16013} leverages an LSTM-based encoder for text-to-Vis dialogues.
Conversely, \textbf{\emph{ncNet}}~\cite{DBLP:journals/tvcg/LuoTLTCQ22} transitions to a Transformer-based model. Its multi-self-attention design eliminates recurrent computations, heightening efficiency. In ncNet, tokenized inputs from three sources are sequenced and merged. Each word is tokenized, masked tokens are populated, and boundary-indicating tokens are added. These tokens undergo vectorization using various embeddings, establishing ncNet as a state-of-the-art in Text-to-Vis, proficiently converting queries into visualization codes.

$\bullet$ \textit{Graph-based Encoder.} As the field of Text-to-Vis progresses, there is a notable shift toward leveraging more complex and efficient encoding methods for input data. Unlike sequence-based methods that linearly process input data, graph-based encoders can capture non-linear relationships within the data, thus offering a richer and more contextually accurate representation of the input. 


A notable work in this direction is \textbf{\emph{RGVisNet}}~\cite{DBLP:conf/kdd/SongZWJ22}. It merges sequence and graph-based encoding in a novel retrieval-generation approach. The input natural language query(NLQ) is parsed to extract relevant VQL from its codebase, achieved by retrieving schemas in the NLQ, performing schema linking, and locating similar VQLs from the codebase. The NLQ is embedded through an LSTM encoder, while the candidate VQLs are processed through a Graph Neural Network (GNN) encoder using an abstract syntax tree (AST) representation. The relevance between NLQ and VQL embeddings is assessed using cosine similarity, with the embeddings then funneled into a Transformer encoder to ascertain relationships and yield the final output.\\

\noindent \textbf{Decoder. } Decoders in Text-to-Vis systems translate encoded textual input into coherent visualizations. Existing approaches have incorporated LSTM, transformer, and grammar-based decoders.

$\bullet$ \textit{Monolithic Decoder.} In the context of Text-to-Vis tasks, monolithic decoders utilize a single, end-to-end model, often based on RNNs, LSTMs, or Transformer architectures, to transform a natural language description into a complete and coherent visual representation by sequentially generating components of a visualization, conditioned on an encoded representation of the input text.

\textbf{\emph{Seq2Vis}}~\cite{DBLP:conf/sigmod/Luo00CLQ21} uses an LSTM decoder within its architecture to generate visual queries. The attention mechanism it incorporates enables dynamic consideration of the input sequence's segments during output generation.
Conversely, \textbf{\emph{ncNet}}~\cite{DBLP:journals/tvcg/LuoTLTCQ22} employs a Transformer-based encoder-decoder approach. Both its encoder and decoder are built using self-attention blocks, optimizing inter-token relationship processing. This design provides flexibility in sequence translation, with the auto-regressive decoder ensuring coherent and logically sequenced outputs.

$\bullet$ \textit{Grammar-based Decoder.}
\textbf{\emph{RGVisNet}}~\cite{DBLP:conf/kdd/SongZWJ22} introduces a grammar-aware decoder tailored for VQL revision. Given VQL's strict and defined grammar, similar to programming languages, leveraging this structure becomes essential. This approach mirrors text-to-SQL tasks, where integrating grammar as inherent knowledge effectively guides code generation. RGVisNet adapts the SemSQL grammar to support DV queries. The core decoder in RGVisNet adapts an LSTM-based structure underpinned by the formation of a context-free grammar tree. As the model traverses this tree, it leverages an LSTM model at every step to opt for the most likely branch, based on prior routes. 

\subsubsection{Foundation Language Model Stage}
Foundation language models (FMs), especially large language models such as CodeX~\cite{DBLP:journals/corr/abs-2107-03374} and GPT-3, have revolutionized natural language processing with their ability to generate contextually accurate text. This is leveraged to advance the field of Text-to-Vis towards a new set of approaches.

$\bullet$ \textit{Zero-Shot Prompting.} Zero-shot prompts refer to the use of untrained prompts to guide LLMs in generating visualization codes straight from textual or spoken queries. Leveraging LLMs' natural language understanding capabilities, zero-shot prompting in text-to-visualization systems employs carefully crafted prompts as guiding instructions, steering the models to generate specific and contextually appropriate visualizations based on user input. Mitra et al.~\cite{DBLP:journals/corr/abs-2207-00189} developed a prototype web application by prompting CodeX. \textbf{\emph{Chat2VIS}}~\cite{DBLP:journals/corr/abs-2302-02094} also chose the model CodeX and specifically included a code prompt component to guide the LLM. These two methods both output visualization specification code directly.

$\bullet$ \textit{Few-Shot Prompting.} Few-shot methods employ limited examples to guide LLMs toward desired outputs. \textbf{\emph{NL2INTERFACE}}~\cite{DBLP:journals/corr/abs-2209-08834} utilizes CodeX by first preparing examples that translate natural language queries into a specific VQL format named SPS. This step forms a suitable prompt for in-context learning by CodeX. Subsequently, given the natural language queries and a database catalog, Codex predicts the corresponding VQL. Finally, NL2INTERFACE maps these SPS representations to generate interactive interfaces, following a procedure similar to PI2 based on a predefined and extensible cost model.

\begin{table*}[htbp]
\caption{Comparative Analysis of Evaluation Metrics\label{tab:metrics}}
\begin{center}
\resizebox{\textwidth}{!}{
\begin{tabular}{llll}
        \toprule
            \textbf{Type} & \textbf{Method} & \textbf{Advantages} & \textbf{Disadvantages} \\
        \midrule
            \multirow{3}{*}{String-based Matching}
              & Exact String Match & High efficiency, wide applicability & Cannot handle alias expressions \\
              & Fuzzy Match & Suitable for complex queries & Insufficient precision \\
              & Component Match & Can handle simple alias expressions & Needs to be customized \\
             \midrule[0.5pt]
             \multirow{2}{*}{Execution-based Matching} 
             & Naive Execution Match & Convenient, robust to alias expressions & Prone to false positives \\
             & Test Suite Match & Can handle semantically close expressions & Needs to be customized\\
             \midrule[0.5pt]
             \multirow{1}{*}{Manual Evaluation} 
             & Manual Evaluation  & Precise, flexible & High cost, low efficiency \\
        \bottomrule
\end{tabular}}
\end{center}
\end{table*}

\begin{tcolorbox}[colback=gray!10,colframe=black!50,title=Takeaways for Approaches]
\begin{itemize}
\item Traditional Stage:
\begin{itemize}
\item Rule-based: Interpretable, limited adaptability. Use for interpretability.
\item Template-based: Fast, struggles with novel queries. Use for well-defined scenarios.
\end{itemize}
\item Neural Network Stage:
\begin{itemize}
\item Encoders: Sequence (flexible, data-hungry), Graph (captures relations, complex). Choose based on data complexity and structure.
\item Decoders: Monolithic (unified, may miss nuances), Skeleton (captures structure, limited generalization), Grammar (aligns with formal languages, limited complexity), Execution (validates output, might be slower). Choose based on generalization needs.
\end{itemize}
\item Foundation Language Model Stage:
\begin{itemize}
\item PLM-based: Strong performance, hard to interpret. Prefer for performance.
\item LLM-based: Strong performance and generalization, requires prompt engineering. Prefer for generalization (with prompt engineering).
\end{itemize}
\end{itemize}
\end{tcolorbox}

\section{Evaluation Metrics}
\label{section:metrics}

Evaluation metrics play a pivotal role in assessing the performance of semantic parsers for both Text-to-SQL and Text-to-Vis tasks. Table~\ref{tab:metrics} provides a comparative analysis of commonly used metrics.

\subsection{Text-to-SQL Metrics}

\subsubsection{String-based Matching}

String-based matching metrics evaluate the textual match between the generated SQL query and the ground truth. 

\noindent \textbf{Exact String Match.} Exact String Match~\cite{DBLP:conf/acl/RadevKZZFRS18} is the strictest form of string-based evaluation, requiring the generated query to be identical to the target query. While efficient and broadly applicable, it can overlook semantically equivalent queries with slight syntactic differences.

\noindent \textbf{Fuzzy Match.} Fuzzy Match allows for approximate matching, quantifying similarity by assigning scores based on string closeness, such as BLEU~\cite{George2002bleu}. It offers flexibility for minor discrepancies but may be overly lenient, potentially overlooking significant errors~\cite{DBLP:conf/lrec/LinWZE18}.

\noindent \textbf{Component Match.} Component matching, such as Exact Set Match~\cite{DBLP:conf/emnlp/YuZYYWLMLYRZR18}, focuses on individual components or segments of the predicted SQL query. It ensures the correctness of each component independently.

\subsubsection{Execution-based Matching}

Execution-based matching evaluates the correctness of a SQL query based on its execution results.

\noindent \textbf{Execution Match.} Execution match considers the generated query correct if its results match the reference query, regardless of syntactic differences. It is beneficial when distinct queries can lead to the same output, avoiding false negatives from string-based metrics.

\noindent \textbf{Test Suite Match.} Test Suite Match, proposed by Zhong et al.~\cite{DBLP:conf/emnlp/ZhongYK20}, creates multiple knowledge base variants to differentiate between predicted and reference queries. A query is considered correct only if its execution results align with the reference across all variants, ensuring consistency.

\subsubsection{Manual Evaluation}

\noindent \textbf{Human Evaluation of SQL.} In the Text-to-SQL task, human evaluation discerns semantic equivalence when execution results differ but are valid in real-world scenarios. Dahl et al.~\cite{DBLP:conf/naacl/DahlBBFHPPRS94} introduced an approach where a result is correct if it falls within a predefined interval which often requires human judgment.

\subsection{Text-to-Vis Metrics}

\subsubsection{String-based Matching}

\noindent \textbf{Exact String Match.} Exact String Match, often referred as Overall Accuracy in the context of Text-to-Vis~\cite{DBLP:conf/kdd/SongZWJ22}, directly measures the matches between the predicted visualization query and the ground truth query. This metric reflects the comprehensive performance of the models.

\noindent \textbf{Component Match.} Component matching, such as in RGVisNet~\cite{DBLP:conf/kdd/SongZWJ22} and Seq2Vis, focuses on individual components of the predicted visualization specification. It ensures the correctness of each component independently.

\subsubsection{Manual Evaluation}

\noindent \textbf{User Study of Vis.} In Text-to-Vis, user studies evaluate models' practical effectiveness, user-friendliness, and efficiency. They capture user feedback on system speed, ease of use, preferences, and suggestions for improvements, offering insights into real-world applicability.

\begin{tcolorbox}[colback=gray!10,colframe=black!50,title=Takeaways for Evaluation Metrics]
\begin{itemize}
\item Text-to-SQL: String-based (Exact, Fuzzy, Component), Execution-based (Execution, Test Suite), Manual (Human Evaluation)
\item Text-to-Vis: String-based (Exact, Component), Manual (User Study)
\item Guidelines: Use string-based metrics for syntactic correctness. Use execution-based metrics for semantic equivalence. Manual evaluation can be considered to capture nuances and subtleties that automated metrics might miss. 
\end{itemize}

\end{tcolorbox}

\subsection{System Architectures}
System architecture is crucial in shaping the capabilities of natural Language interfaces for tabular data querying and visualization. Various architectural paradigms have emerged as the field has evolved, each tailored to specific challenges and needs. While in-depth analyses and comparisons of earlier systems can be found in surveys like~\cite{DBLP:conf/sigmod/GkiniBKI21,DBLP:journals/pvldb/KimSHL20,DBLP:journals/vldb/AffolterSB19}, this section will categorize these systems into four main architectural types: rule-based systems, parsing-based systems, multi-stage systems, and end-to-end systems. Table~\ref{tab:systems} presents a comprehensive overview of various Text-to-SQL and Text-to-Vis systems.

\subsubsection{Rule-based System}
Rule-based systems stand as foundational architectures for natural language interfaces to databases. These systems leverage a set of predefined rules, mapping natural language inputs directly to database queries or visualizations. For Text-to-SQL, systems like PRECISE~\cite{DBLP:conf/coling/PopescuAEKY04} and NaLIR~\cite{DBLP:journals/pvldb/LiJ14} employ rule-based strategies, translating linguistic patterns into SQL queries. In the Text-to-Vis context, DataTone~\cite{DBLP:conf/uist/GaoDALK15} represents this approach, converting user language into visualization specifications via established patterns. While precise, rule-based systems can face challenges in scalability and adaptability to diverse linguistic constructs.

\subsubsection{Parsing-based System}
Parsing-based systems primarily focus on understanding the inherent grammatical structure of the input question. Drawing inspiration from traditional linguistic parsing, these systems convert natural language questions into syntactic structures or logical forms. In the field of Text-to-SQL, systems such as SQLova~\cite{DBLP:journals/corr/abs-1902-01069} and Seq2Tree~\cite{DBLP:conf/acl/DongL16} utilize semantic parsers to bridge the gap between natural language and structured database queries. For Text-to-Vis, systems ncNet~\cite{DBLP:journals/tvcg/LuoTLTCQ22} process user queries through semantic parsing, transforming them into Visualization Query Languages (VQL). Parsing-based systems prioritize linguistic structure and semantics, offering depth in understanding, but might struggle with the variability and ambiguity inherent to natural language.

\subsubsection{Multi-stage System}
Multi-stage systems in natural language interfaces for tabular data operate through sequenced processing pipelines. These systems dissect the overarching task into distinct stages, each addressing a particular sub-task. This layered approach allows for focused improvements at every juncture. Within the Text-to-SQL domain, the DIN-SQL system~\cite{DBLP:journals/corr/abs-2304-11015} exemplifies this architecture, segmenting SQL generation into stages for schema linking, query classification and decomposition, SQL generation, and self-correction.
In the Text-to-Vis sphere, DeepEye~\cite{DBLP:conf/sigmod/LuoQ00W18} emerges as a notable multi-stage system to discern the quality of visualizations, rank them, and optimally select the top-k visualizations from a dataset.
By segmenting the process, multi-stage systems can apply tailored techniques to each segment, enhancing accuracy. However, the modular approach demands careful orchestration between stages to ensure coherency in the final output and can potentially bring higher computational cost.

\subsubsection{End-to-end System}
End-to-end systems represent a holistic approach to natural language interfaces for tabular data. Rather than relying on intermediate representations or multi-phase processing, these systems process input questions and directly generate the desired output in one cohesive step. 
For example, Photon~\cite{DBLP:conf/acl/ZengLHSXLK20} offers a modular framework tailored for industrial applications of Text-to-SQL systems. It takes a user's question and a database schema, directly generating SQL and executing it to produce the desired result, with its core strength lying in its SQL parser and a unique confusion detection mechanism. Another exemplar is VoiceQuerySystem~\cite{DBLP:conf/sigmod/SongWZJ22}, which elevates the user experience by converting voice-based queries directly into SQL, bypassing the need for text as an intermediary.
Similarly, in the Text-to-Vis domain, Sevi~\cite{DBLP:conf/sigmod/TangLOLC22} stands out as an end-to-end visualization assistant. It empowers novices to craft visualizations using either natural language or voice commands. Furthermore, DeepTrack~\cite{DBLP:journals/pvldb/LuoLZYZ0020} integrates data preparation, visualization selection, and intuitive interactions within a singular framework, exemplifying the comprehensive capabilities of end-to-end systems. 

\begin{table*}[htbp]
\caption{Comparison of Different Systems of Text-to-SQL and Text-to-Vis\label{tab:systems}}
\begin{center}
\resizebox{\textwidth}{!}{
\begin{tabular}{lllll}
\toprule
\textbf{Type} & \textbf{SQL system example} & \textbf{Vis system example} & \textbf{Advantages} & \textbf{Disadvantages} \\
\midrule
Rule-based & NaLIR, PRECISE & DataTone & Robustness and consistency for familiar queries & Limited adaptability \\
\midrule
Parsing-based & SQLova, Seq2Tree & ncNet & Grasps deeper language structures & Struggles with ambiguity \\
\midrule
Multi-stage & DIN-SQL & DeepEye  & Enhanced accuracy and flexibility & Synchronization challenges  \\
\midrule
End-to-end & Photon, VoiceQuerySystem & Sevi, DeepTrack  & High adaptability, unified training process & Difficult to interpret and debug \\
\bottomrule
\end{tabular}}
\end{center}
\end{table*}

\subsection{User-centric Analysis}

When designing natural language interfaces for tabular data querying and visualization, it is essential to consider the needs of different user groups.

\subsubsection{Basic Users}

For basic users with limited technical backgrounds, rule-based systems are most suitable due to their simplicity and accuracy in well-defined domains. End-to-end systems are recommended for those needing flexibility to handle diverse queries effortlessly. Users with stronger technical skills or those working with complex data structures may prefer parsing-based systems, which excel in handling intricate linguistic structures.

\subsubsection{Professional Users}
Professional users in corporations and academic institutions often deal with large volumes of data and have higher querying frequencies. In stable and standardized data environments, rule-based systems ensure reliable performance for repetitive queries. For complex data environments requiring integration and analysis of different data types, multi-stage systems offer enhanced adaptability and accuracy. In fast-paced environments where speed and efficiency are crucial, end-to-end systems are the most suitable choice, minimizing latency and allowing for rapid adaptation to changing data landscapes.

\begin{tcolorbox}[colback=gray!10,colframe=black!50,title=Takeaways for System Design]
\begin{itemize}
\item Rule-based: Precise but limited adaptability. Suitable for well-defined domains. Recommended for basic users seeking simplicity and professional users in stable environments.
\item Parsing-based: Focuses on grammar but struggles with ambiguity. Ideal for complex linguistics.  Recommended for basic users with technical expertise.
\item Multi-stage: Sequential processing for enhanced accuracy. Best for complex data. Recommended for professional users dealing with complex data.
\item End-to-end: Holistic and highly adaptable. Perfect for fast-paced environments.Recommended for basic users requiring flexibility and professional users prioritizing efficiency.
\end{itemize}

\end{tcolorbox}

\section{Future Research Directions}
\label{section:future}

\begin{table*}[htbp]
\caption{Comparison between Text-to-SQL and Text-to-Vis Research\label{tab:future}}
\begin{center}
\resizebox{\textwidth}{!}{
\begin{tabular}{lll}
\toprule 
\textbf{Aspects} & \textbf{Text-to-SQL} & \textbf{Text-to-Vis} \\
\midrule
\textbf{Neural Models and Approaches} & Notable advancements with a variety of models & Still in nascent stages with limited models \\
\midrule
\textbf{Integration of LLMs} & Preliminary efforts with room for more exploration & Limited work, significant potential \\
\midrule
\textbf{Learning Methods} & Mainly supervised; early exploration of semi-supervised methods & Predominantly supervised \\
\midrule
\textbf{Datasets} & Several large-scale datasets available & Fewer datasets; need for more diversity \\
\midrule
\textbf{Robustness and Generalizability} & Increased focus, especially for complex queries & Emerging focus; essential for diverse visualizations \\
\midrule
\textbf{Advanced Applications} & Integration with chatbots, recommendation systems & Potential for multimodal systems, dynamic visualizations \\
\bottomrule
\end{tabular}}
\end{center}
\end{table*}


As natural language interfaces for tabular data querying and visualization evolve, new challenges and opportunities emerge. This section highlights six pivotal areas that promise to shape the domain's future, emphasizing the ongoing research evolution and its potential. Table~\ref{tab:future} compares Text-to-SQL and Text-to-Vis tasks across these research directions.

\subsection{Advancing Neural Models and Approaches}
The landscape of Natural Language Interfaces for Tabular Data has seen impressive strides, especially with the advent of neural models in the text-to-SQL domain. However, there remains substantial room for improvement and innovation. While plenty of models have been proposed for text-to-SQL tasks, continual refinement is essential to handle more complex queries, multi-turn interactions, and domain-specific problems~\cite{DBLP:conf/coling/Deng0022}.
Concurrently, the text-to-visualization domain hasn't witnessed the same influx of neural network-based models. The challenges here are multifold: from generating diverse visualizations based on user intent to ensuring those visualizations maintain both accuracy and aesthetic appeal~\cite{DBLP:journals/tvcg/ShenSLYHZTW23}.
For both domains, it is vital to push the boundaries of current neural architectures. This could involve exploring deeper networks, advanced attention mechanisms, or hybrid models combining rule-based logic with neural insights. Leveraging external knowledge bases, transfer learning, and multi-modal strategies could further optimize the interpretation and translation of user intent into SQL queries or visual representations.

\subsection{Harnessing Potential of Large Language Models}
Large Language Models (LLMs) like ChatGPT have revolutionized various Natural Language Processing domains with their profound text understanding and generation capabilities. Despite this, exploring LLMs in the context of natural language interfaces for databases remains relatively nascent.
While preliminary efforts have begun integrating LLMs into text-to-SQL and text-to-visualization systems~\cite{DBLP:journals/corr/abs-2304-11015,DBLP:journals/corr/abs-2209-08834}, the vast potential of LLMs has not been fully harnessed. Their ability to capture context, understand nuances, and generalize from limited examples could be invaluable in understanding and translating complex user queries.
However, merely deploying LLMs without customization might not be optimal. Future research should focus on tailoring these models to the specific challenges of querying and visualization. This might involve adapting LLMs on domain-specific datasets, integrating them with existing architectures, or developing novel prompting strategies to better align them with the tasks at hand.

\subsection{Exploring Advanced Learning Methods}
The heavy reliance on traditional supervised learning on large labeled datasets poses challenges for evolving natural language interfaces for tabular data. This underscores the need for alternative learning approaches. Semi-supervised and weakly supervised methods, which capitalize on unlabeled data or weak supervision signals, present viable solutions~\cite{DBLP:conf/emnlp/GuoLLZ21}. For example, implicit user interactions might offer weak guidance for model refinement.
Additionally, parameter-efficient training methods like LoRA~\cite{DBLP:conf/iclr/HuSWALWWC22} have demonstrated superior data efficiency, especially in low-resource settings, compared to traditional fine-tuning methods. Fusing large pre-trained models with these parameter-efficient techniques hints at a promising future for data-efficient semantic parsing.  

\subsection{Constructing Large-Scale and Diverse Datasets}
The potency of natural language interfaces for databases depends on high-quality, diverse datasets. 
While several datasets are tailored for text-to-SQL and text-to-vis tasks, there's a pressing need for even larger-scale, more varied datasets. Such datasets foster better generalization and robustness to a broad spectrum of user queries, spanning various domains and complexities.
Moreover, the current dataset landscape is predominantly English-centric, overlooking the global spectrum of data user~\cite{DBLP:conf/acl/GuoSWLFLYL20}. Embracing multilingual or under-represented language datasets can amplify the reach and inclusivity of these interfaces.

\subsection{Advancing Robustness and Generalizability}
As natural language interfaces for tabular data become more integral in various applications, the robustness and generalizability of the underlying models and systems are central. It's not just about achieving high performance on benchmark datasets; real-world scenarios demand models that can reliably handle diverse, unexpected, and sometimes adversarial inputs.

$\bullet$ \textit{Robustness Against Adversarial and Out-of-Distribution Perturbations.} As with many machine learning models, adversarial attacks or unexpected inputs can pose significant challenges. There's a need for models that can gracefully handle and respond to such inputs without compromising on accuracy or reliability. This involves developing models inherently resistant to such perturbations and creating datasets that can effectively test such robustness~\cite{DBLP:conf/iclr/Chang0DPZLLZJLA23}.

$\bullet$ \textit{Compositional Generalization.} The ability for models to understand and combine known concepts in novel ways is vital. For instance, if a model understands two separate queries, it should ideally be able to handle a composite query that combines elements of both. This capability ensures that models can effectively tackle unseen queries by leveraging their understanding of underlying concepts.

$\bullet$ \textit{Domain Generalization.} As these interfaces permeate various sectors, models should adapt across domains and incorporate domain-specific knowledge. This ensures that, while retaining versatility, models are attuned to the nuances of diverse queries, from finance to healthcare and beyond~\cite{DBLP:conf/emnlp/GanCP21}. 

\subsection{Pioneering Advanced Applications in the LLM Era}

The LLM era presents opportunities for revolutionizing applications and systems of natural language interfaces for databases. 

$\bullet$ \textit{Multimodal Systems.} Combining the power of LLMs with other modalities, such as visual or auditory inputs, can lead to the creation of truly multi-modal systems. Imagine querying a database not just with text, but with images, voice commands, or even gestures. Such systems can cater to a broader audience and offer more dynamic and natural interactions. 

$\bullet$ \textit{Integrated Systems.} As LLMs continue to excel in various tasks, there's potential for integrating natural language interfaces with other functionalities, like document summarization, recommendation systems, or even chatbots. This can result in comprehensive systems where users can query data, get summaries, seek recommendations, and more, all within a unified, language-centric interface.

$\bullet$ \textit{User-Centric Design.}  The LLM era emphasizes user interaction. There's a need for applications prioritizing user experience, offering intuitive interfaces, interactive feedback, and personalized responses. By harnessing the capabilities of these models and focusing on creating holistic, user-centric applications, we can set the stage for a future where data interaction is both efficient and delightful. 

\section{Conclusion}
\label{section:conclusion}

In this survey, we explore Natural Language Interfaces for Tabular Data Querying and Visualization in-depth, delving into the intricacies of the field, its evolution, and the challenges it addresses. We trace its evolution from foundational problem definitions to state-of-the-art approaches. We highlight the significance of diverse datasets fueling these interfaces and discuss the metrics that gauge their efficacy. By exploring system architectures, we examine the differences of distinct system designs. Lastly, our gaze turns toward the horizon, pointing to promising research avenues in the era of Large Language Models. As this dynamic field evolves, our exploration offers a concise snapshot of its current state, challenges, and potential.\\

\smallskip
\noindent\textbf{Acknowledgements.} We are grateful to the anonymous reviewers for their constructive comments on this paper. 
The research of Victor Junqiu Wei was supported in part by the HKUST-WeBank Joint Laboratory Project (Ref. Code: HWJL-2023.003, Project No.: WEB24EG01-A). 
\bibliographystyle{abbrv}
\bibliography{refabbr_simp}

%

\begin{IEEEbiography}
[{\includegraphics[
width=1in,height=1.25in,
clip,keepaspectratio]{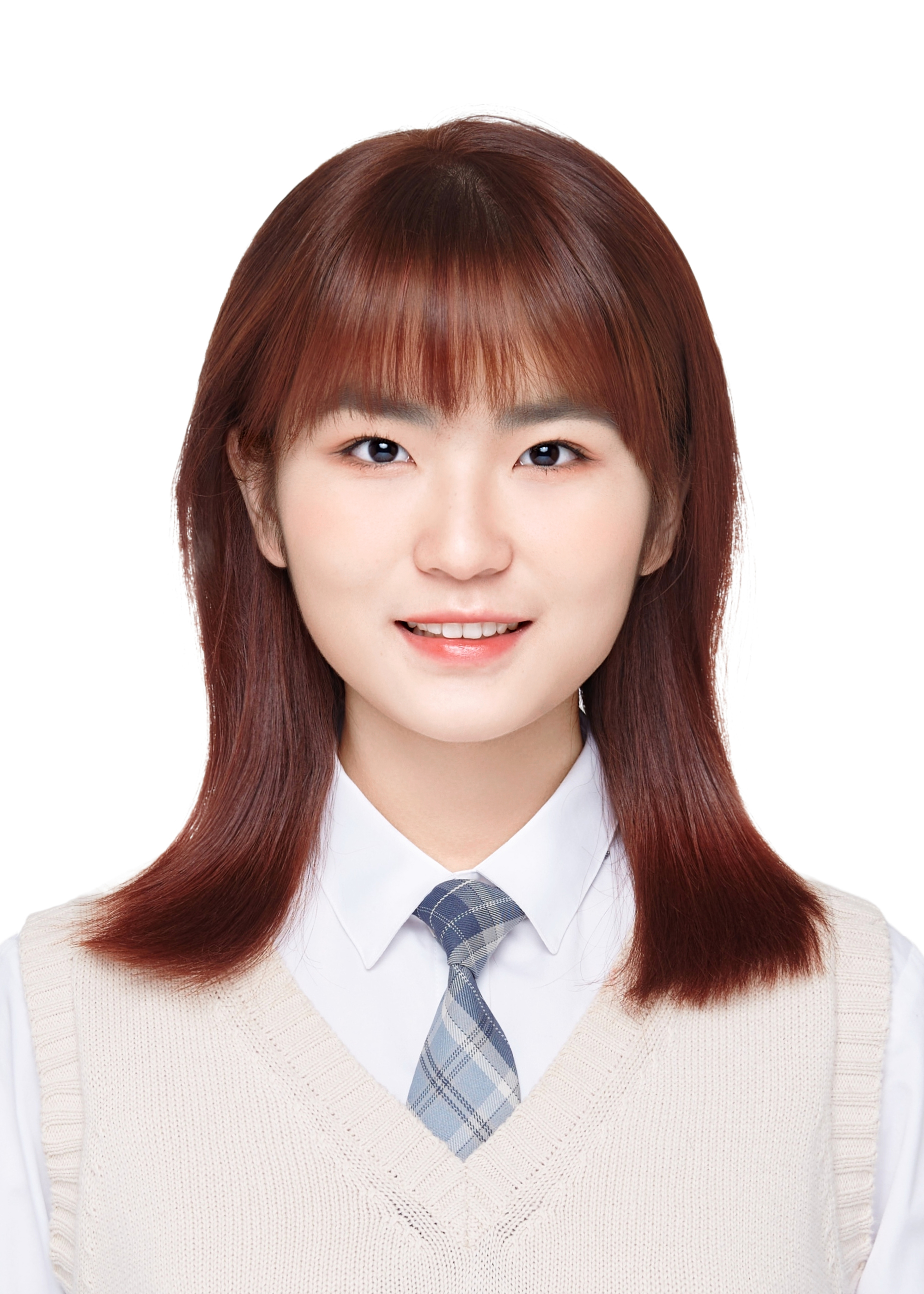}}]{Weixu Zhang} received the B.Eng. degree from Xi'an Jiaotong University, China and Engineer's degree from Paris-Saclay University, France. She is currently a master candidate in Xi'an Jiaotong University. Her research interests include semantic parsing, data mining, and natural language processing.
\end{IEEEbiography}

\vspace{-25pt}


\begin{IEEEbiography}
[{\includegraphics[
width=1in,height=1.25in,
clip,keepaspectratio]{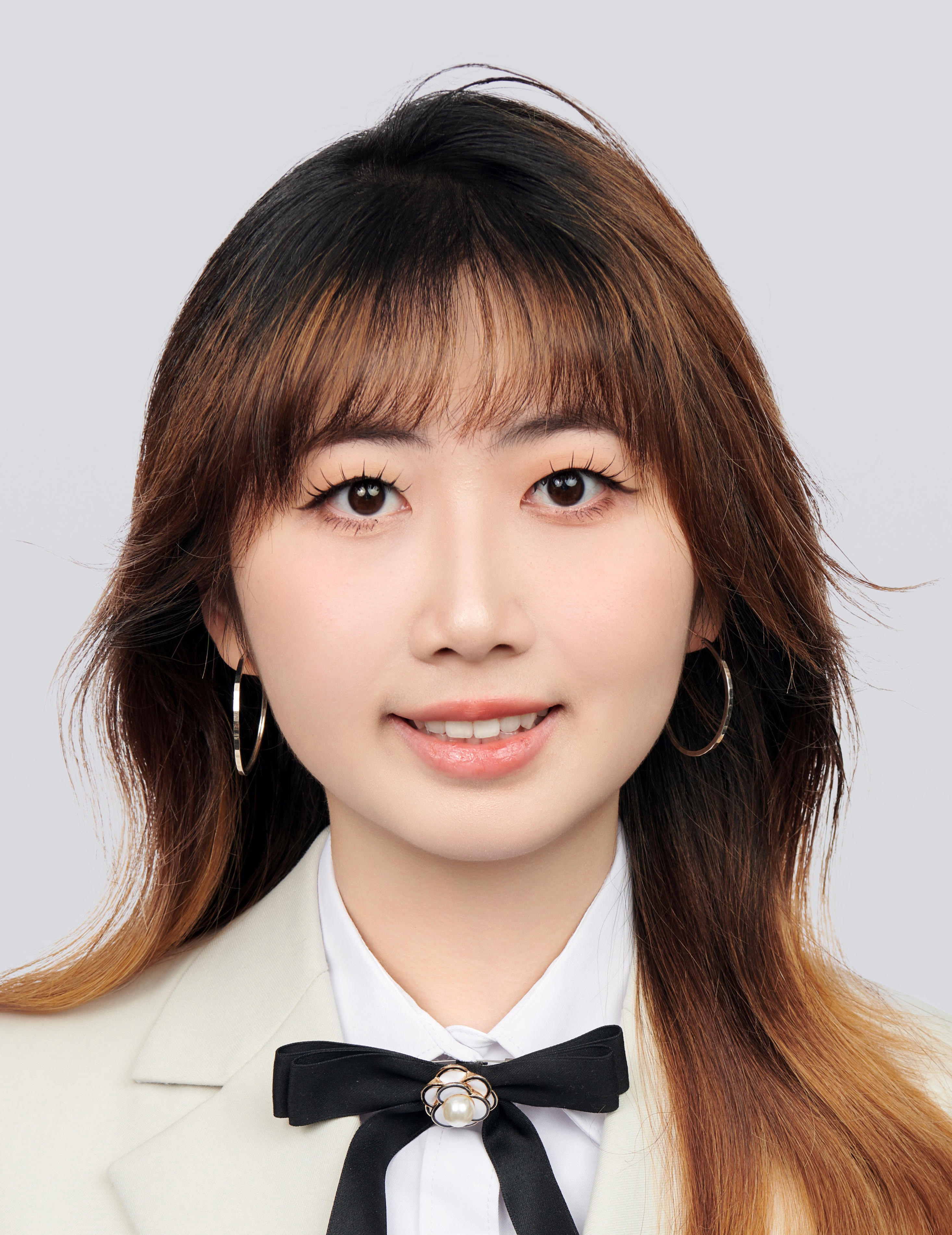}}]{Yifei Wang} is currently pursuing a Bachelor of Applied Science in Machine Intelligence at the University of Toronto. Her research interests include natural language processing, data visualization, and data querying.  
\end{IEEEbiography}

\vspace{-25pt}


\begin{IEEEbiography}[{{\includegraphics[
width=1in,height=1.25in,
clip,keepaspectratio]{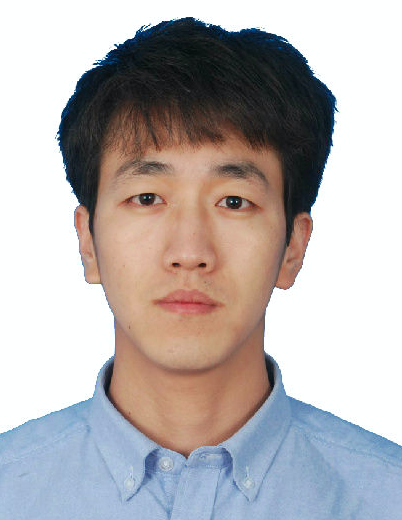}}}]{Yuanfeng Song} received the M.Phil degree in computer science from the Hong Kong University of Science and Technology in 2012. He is currently a senior researcher at WeBank AI. His research interests include information retrieval, natural language processing, and speech recognition.
\end{IEEEbiography}

\vspace{-25pt}


\begin{IEEEbiography}
    [{\includegraphics[width=1in,height=1.25in,
    clip,keepaspectratio]{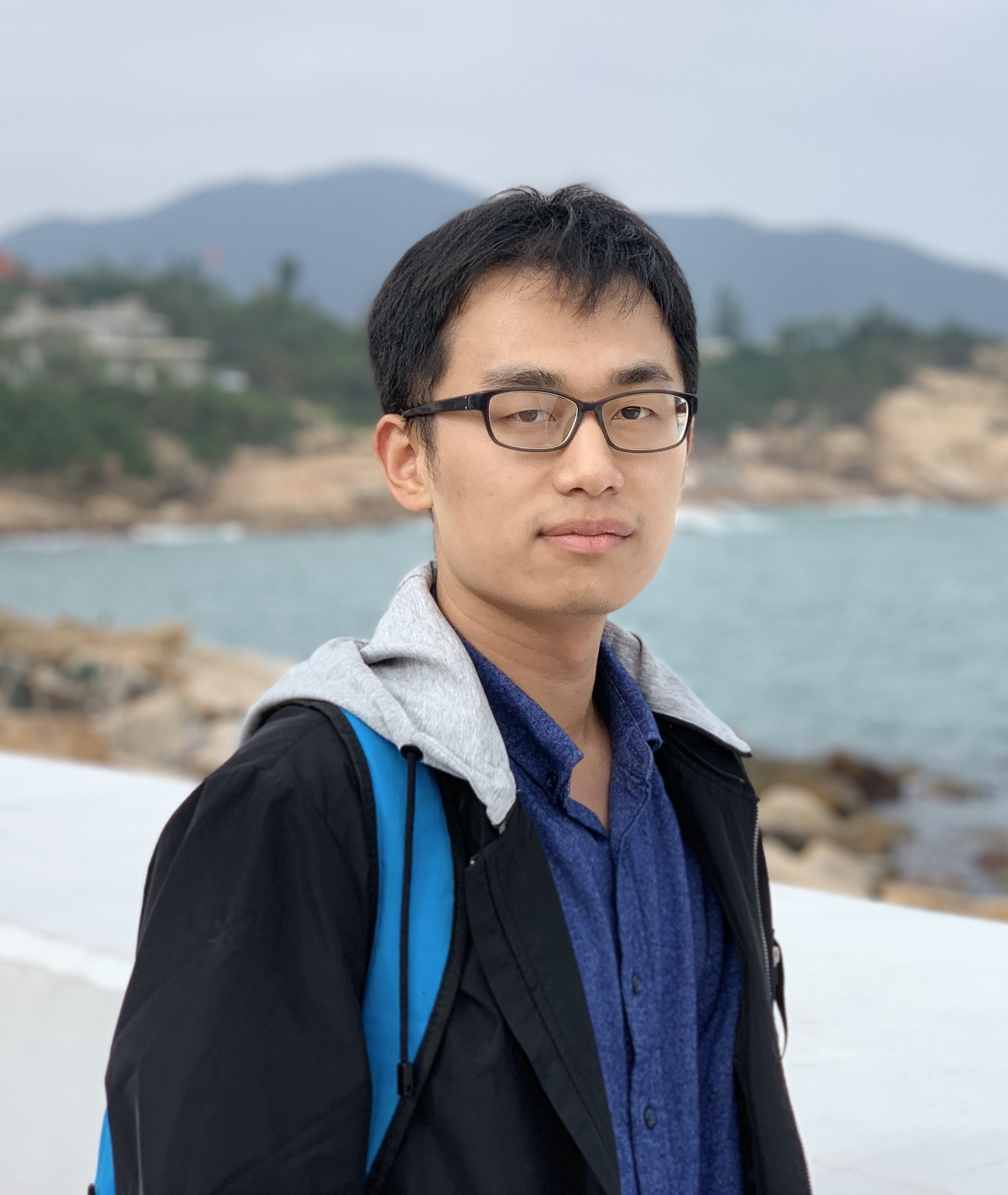}}]{Victor Junqiu Wei}
      is currently working as a research assistant professor in the Department of Computer Science and Engineering (CSE), the Hong Kong University of Science and Technology (HKUST). He obtained his bachelor degree from Nanjing University and PhD degree from Department of Computer Science and Engineering, the Hong Kong University of Science and Technology. He also has several years' working experience in the world-famous research labs in the AI industry including Baidu natural language processing (NLP) group, the AI group of WeBank, and Noah's Ark Lab of Huawei. 
  \end{IEEEbiography}

\vspace{-25pt}


\begin{IEEEbiography}
[{\includegraphics[
width=1in,height=1.25in,
clip,keepaspectratio]{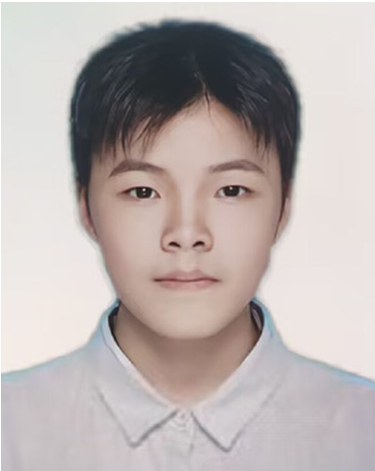}}]{Yuxing Tian} is currently pursuing the B.Eng. degree at Xidian University. His research interests include graph neural networks, large language models, and federated learning.
\end{IEEEbiography}

\vspace{-25pt}


\begin{IEEEbiography}
[{\includegraphics[
width=1in,height=1.25in,
clip,keepaspectratio]{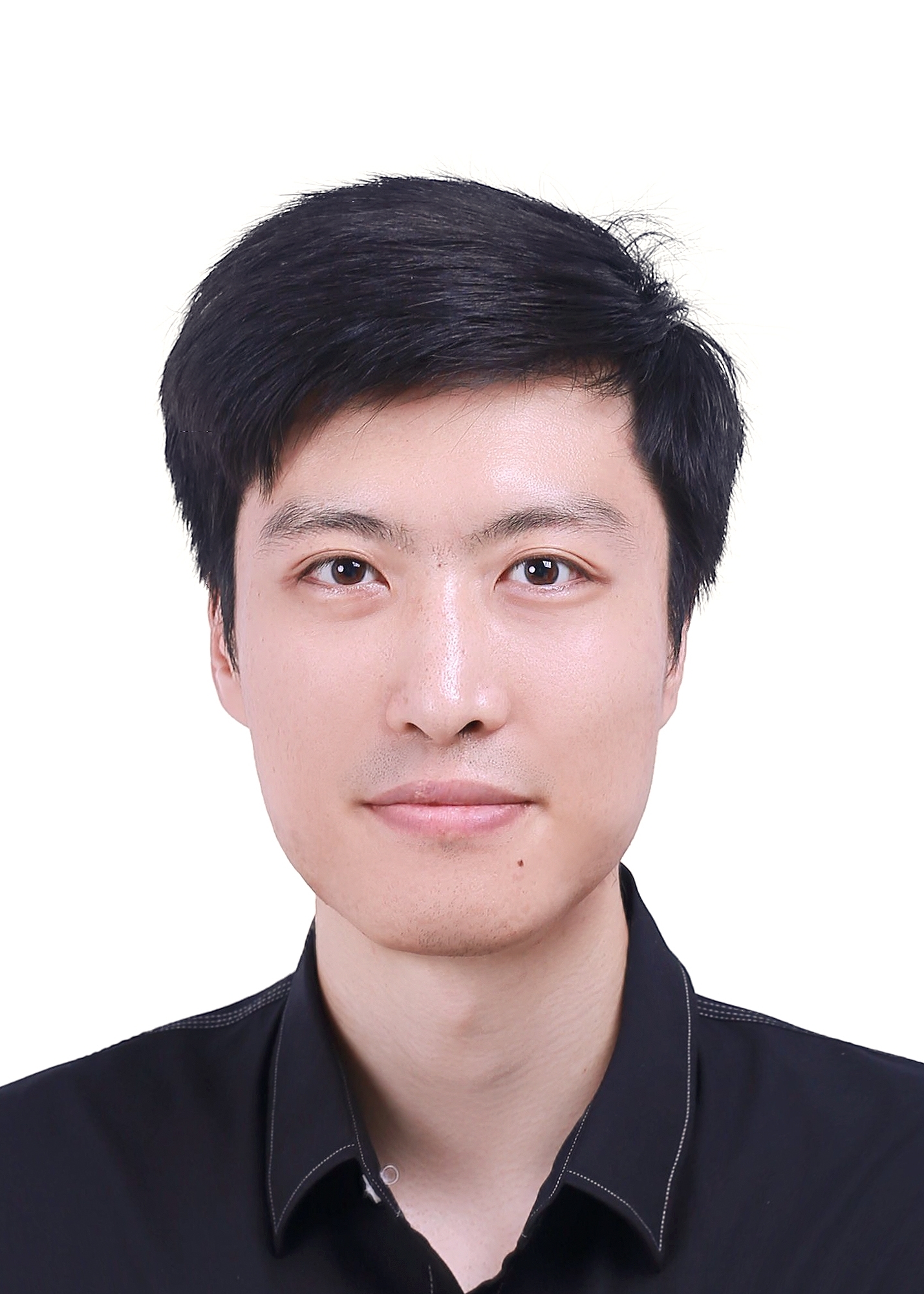}}]{Yiyan Qi} received the B.S in automation engineering and the Ph.D. degree in automatic control from Xi'an Jiaotong University, Xi'an, China, in 2014 and 2021 respectively.  He is currently a Researcher at IDEA. Prior to joining IDEA, he was working in Tencent. His current research interests include abnormal detection, graph mining and embedding, and recommender systems.
\end{IEEEbiography}

\vspace{-25pt}

\begin{IEEEbiography}[{\includegraphics[
width=1in,height=1.25in,
clip,keepaspectratio]{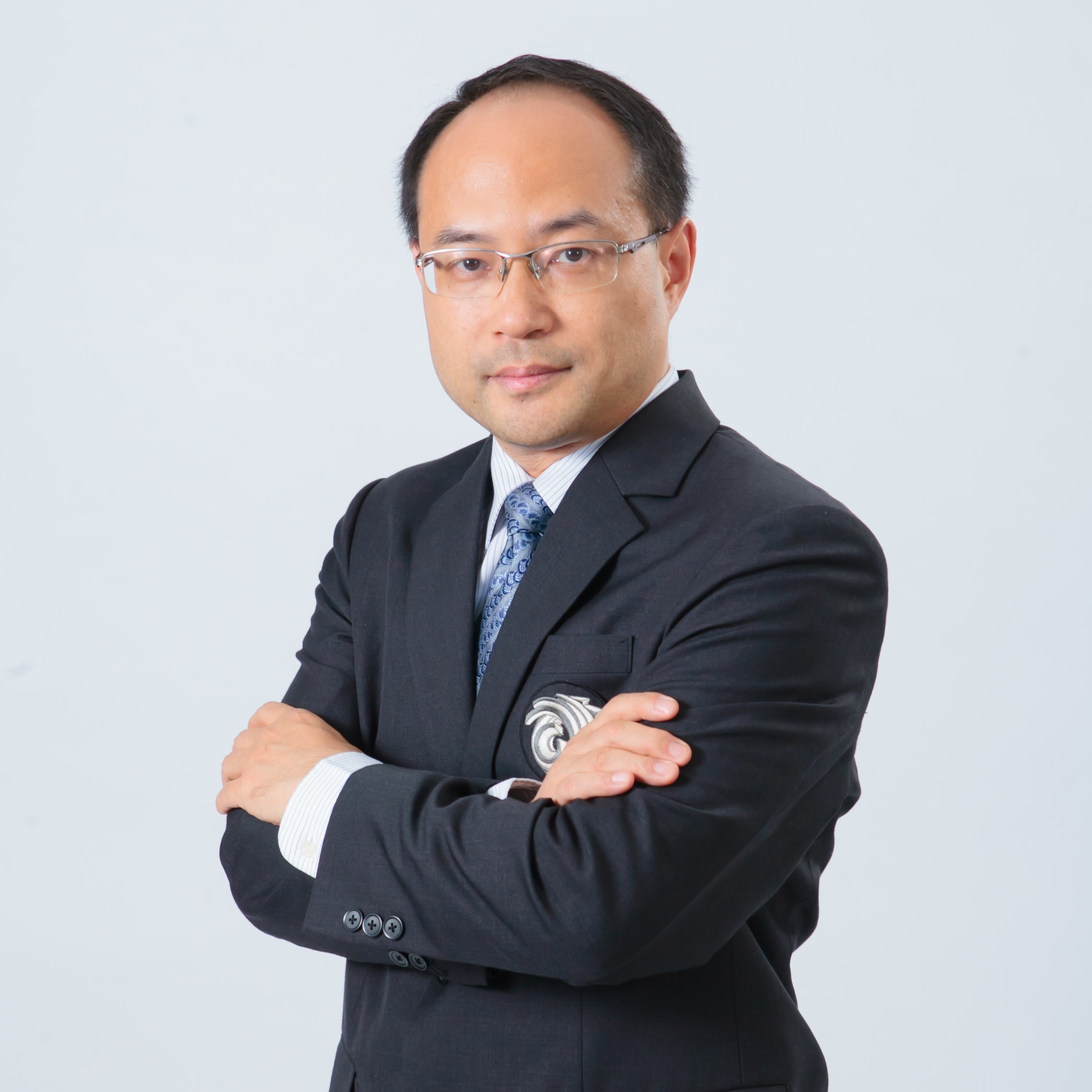}}]{Jonathan H. Chan} is an Associate Professor of Computer Science at the School of Information Technology, King Mongkut's University of Technology Thonburi (KMUTT), Thailand. Jonathan holds a B.A.Sc., M.A.Sc., and Ph.D. degree from the University of Toronto and was a visiting professor back there on several occasions. He also holds a status as a visiting scientist at The Centre for Applied Genomics at Sick Kids Hospital in Toronto. Jonathan is the Director of the Innovative Cognitive Computing (IC2) Research Center at KMUTT. In addition, he is a founding member and a past Chair of the IEEE-CIS Thailand Chapter, the current President (2023-2024) of the Asia Pacific Neural Network Society (APNNS), and a senior member of IEEE, ACM, INNS, and APNNS. His research interests include intelligent systems, cognitive computing, biomedical informatics, data science, and machine learning in general.
\end{IEEEbiography}

\vspace{-25pt}

\begin{IEEEbiography}[{\includegraphics[width=1in,height=1.25in,clip,keepaspectratio]{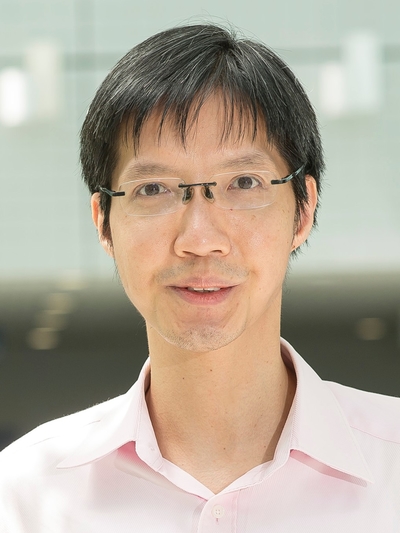}}]{Raymond Chi-Wing Wong}
            is a Professor in Computer Science and Engineering (CSE) of The Hong Kong University of Science and Technology (HKUST). He is currently the associate head of Department of Computer Science and Engineering (CSE). He was the director of the Risk Management and Business Intelligence (RMBI) program (from 2017 to 2019) and the Computer Engineering (CPEG) program (from 2014 to 2016).
            He received the BSc, MPhil and PhD degrees in Computer Science and Engineering in the Chinese University of Hong Kong (CUHK) in 2002, 2004 and 2008, respectively.
        \end{IEEEbiography}
\vspace{-25pt}


\begin{IEEEbiography}[{\includegraphics[
width=1in,height=1.25in,
clip,keepaspectratio]{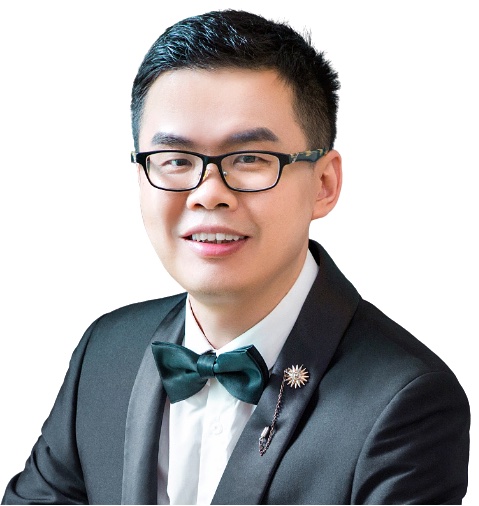}}]{Haiqin Yang} (M’11, SM'18) received the BSc degree in computer science from Nanjing University, Nanjing, China, and the MPhil and PhD degrees from Department of Computer Science and Engineering, The Chinese University of Hong Kong, Hong Kong.  He is currently a principal researcher at IDEA.  His research interests include machine learning, natural language processing, and large language models.  He received the Young Researcher Award of the Asia Pacific Neural Network Society in 2018 and was recognized by AMiner's Most Influential Scholar Award Honorable Mention to the field of AAAI/IJCAI three times. 
\end{IEEEbiography}

\vspace{-25pt}





\end{document}